\documentclass[letterpaper]{article} 
\usepackage{aaai2027}  
\usepackage[hyphens]{url}  
\usepackage{graphicx} 
\usepackage{natbib}  
\usepackage{caption} 
\usepackage{algorithm}
\usepackage{algorithmic}
\usepackage{amsmath} 
\usepackage{amssymb} 
\usepackage{pifont} 
\usepackage[table]{xcolor} 
\usepackage{amssymb}
\usepackage{amssymb}
\usepackage{multirow}
\usepackage{chngcntr}

\definecolor{bgblue}{HTML}{EBF2FA}
\definecolor{bggreen}{HTML}{EBF5EB}
\definecolor{bgpink}{HTML}{FFB6C1}
\usepackage{newfloat}
\usepackage{listings}
\DeclareCaptionStyle{ruled}{labelfont=normalfont,labelsep=colon,strut=off} 
\floatstyle{ruled}
\newfloat{listing}{tb}{lst}{}
\floatname{listing}{Listing}

\usepackage{booktabs}
\usepackage{url}

\title{LeapBot-WA: World-Anchor Action Models via Predictive Latent Alignments}
\author{
    Pei Liu$^{1,2}$\equalcontrib, Nan Zheng$^{3}$\equalcontrib, Lang Zhang$^{3}$\equalcontrib\thanks{Project leader.}, Daojie PENG$^{1}$, Yanan Zhang$^{3}$, Feilong Kong$^{4}$, Mingyue Feng$^{3}$, Jiachao Liu$^{3}$, Yaonong Wang$^{3}$, Qifeng Chen$^{2}$, Jun Ma$^{1,2}$\corresponding
}
\affiliations{
    \textsuperscript{\rm 1}The Hong Kong University of Science and Technology (Guangzhou)\\
    \textsuperscript{\rm 2}The Hong Kong University of Science and Technology\\
    \textsuperscript{\rm 3}Leapmotor\\
    \textsuperscript{\rm 4}Southeast University\\

}

\begin{document}

\maketitle

\begin{abstract}

World Action Models (WAMs) have emerged as a powerful paradigm for embodied intelligence, yet the prevailing reliance on pixel-level video generation creates a fundamental bottleneck. Forcing models to reconstruct task-irrelevant visual details dissipates representational capacity and renders policies vulnerable to visual distractors. In this paper, we propose LeapBot-WA, which establishes a novel Predictive-Latent paradigm for WAMs by operationalizing the Joint-Embedding Predictive Architecture (JEPA) as a World-Anchor. Departing from the traditional reliance on visual synthesis, LeapBot-WA shifts the core of world modeling to Predictive Semantic Alignment, extracting abstract physical dynamics directly within a latent foundation space. To bridge the modality gap between non-Gaussian predictive features and diffusion priors, we introduce the Isotropic Semantic Autoencoder (ISAE), which reshapes the anchor’s latent space into a diffusion-friendly manifold to prevent off-manifold drift. Furthermore, we design an Asymmetric Mixture-of-Transformers (MoT) architecture. During training, an Anchor Diffusion Transformer acts as a privileged dynamics expert to guide the Action Diffusion Transformer; at inference, this heavy dynamics branch is pruned, enabling zero-overhead execution. LeapBot-WA achieves state-of-the-art performance among predictive models on LIBERO and matches top-tier generative WAMs on RoboTwin 2.0 without requiring large-scale trajectory pre-training. It further demonstrates superior zero-shot robustness to unseen environments and successful real-world transfer, establishing a highly efficient and robust latent-centric paradigm for scalable robotic control. Code: \url{https://github.com/LeapWM/leapbot-wa}.
\end{abstract}



\section{Introduction}

A cornerstone of general-purpose embodied intelligence is the ability to anticipate how the physical environment evolves under an agent’s actions. This predictive capability has driven a growing line of research on world models (WMs), which learn predictive representations of environment dynamics to support planning and control~\cite{lecun2022path, hafner2023mastering}. In robotic manipulation, this idea has recently evolved into world action models (WAMs), which couple predictive world modeling with action generation in a single framework~\cite{li2025comprehensive, ye2026world}. By learning from videos and robot trajectories, these models offer a promising path toward scalable policy learning with richer physical priors than purely reactive imitation.

Most existing WAMs, however, remain rooted in a design choice inherited from video generation: they learn dynamics by predicting future visual observations. Whether through explicit future-frame synthesis or latent codes optimized to reconstruct video, the dominant training paradigm still treats world modeling as a pixel-level rendering problem~\cite{ wang2024worlddreamer, wang2026world}. This generative approach is fundamentally misaligned with the core objective of manipulation. To act well, an agent does not need to reproduce textures, shadows, lighting, or background appearance; it needs abstract knowledge of how the world changes under intervention~\cite{hansen2024td, bardes2024revisiting}. Forcing models to allocate massive representational capacity to task-irrelevant high-frequency visual details entangles physical dynamics with visual idiosyncrasies. Consequently, policies learned in this way suffer from severe representation waste and exhibit catastrophic vulnerability to visual distractors, struggling to generalize across novel environments~\cite{gupta2024essential, yang2025invariance}.

Recent efforts have begun to relax this dependence on explicit rendering by moving world-action modeling into compressed latent spaces~\cite{yuan2026fast, luo2026being, chen2026abot}. Yet, they reveal a deeper unresolved challenge: \emph{what representation space should a world action model be trained in?} Moving from pixels to latent codes reduces the burden of reconstruction, but compactness alone does not guarantee semantic abstraction or alignment with action-relevant dynamics~\cite{majumdar2023we}. Furthermore, directly injecting off-the-shelf predictive representations into diffusion-based world models introduces a severe distribution mismatch~\cite{zhang2025both}. Predictive features are highly structured and typically non-Gaussian, whereas diffusion models require isotropic priors to ensure stable flow matching. This modality gap often causes off-manifold drift, destabilizing the denoising trajectory~\cite{esser2024scaling}. Finally, existing frameworks tightly couple world-modeling with policy execution. This architectural entanglement forces the policy to rely on the generative branch during deployment, leaving the agent highly vulnerable to visual distractors and domain shifts~\cite{stone2021distracting, hansen2021generalization}.

To overcome these limitations, we propose LeapBot-WA, a predictive-latent framework built on the insight that scalable world-action modeling requires shifting the focus from visual rendering to Predictive Semantic Alignment through a tripartite strategy involving semantic representation, generative distribution, and inference execution. First, to align semantic representations, we establish a Predictive Anchor framework by leveraging the Joint-Embedding Predictive Architecture (JEPA) as a large-scale predictive foundation model to extract abstract physical dynamics from robotic trajectories, ensuring the policy focuses on underlying state transitions rather than task-irrelevant appearance. Second, to align generative distributions, we introduce the Isotropic Semantic Autoencoder (ISAE) to bridge the modality gap between structured, non-Gaussian predictive features and isotropic diffusion priors, thereby preventing off-manifold drift during joint flow matching. Third, to align inference execution, we design an Asymmetric Mixture-of-Tokens (MoT) architecture where an intent-conditioned Anchor Diffusion Transformer (DiT) guides the Action DiT during training. This design elegantly decouples world-model imagination from execution, allowing the heavy-dynamics branch to be omitted during inference for zero-overhead action generation. Our evaluation across diverse robotic manipulation benchmarks and challenging visual settings demonstrates that high-quality action policies do not require pixel-level world generation. LeapBot-WA achieves state-of-the-art performance among predictive models on LIBERO and matches the success rates of top-tier generative WAMs on RoboTwin 2.0, notably without requiring large-scale trajectory pre-training. Beyond benchmark superiority, it demonstrates exceptional zero-shot robustness to unseen environments and visual distractors, while facilitating seamless real-world transfer. These results establish that highly efficient and robust embodied intelligence can be achieved by prioritizing latent-space dynamics over pixel-level synthesis.
Our main contributions are summarized as follows:
\begin{itemize}
    \item \textbf{Predictive-Latent Semantic Alignment:} We propose LeapBot-WA, a paradigm shift that circumvents the reliance on pixel-level video pre-training. By establishing a Predictive Anchor, we successfully transfer and adapt universal physical priors from predictive foundation models to robotic control, extracting abstract dynamics without the burden of visual reconstruction.
    
    \item \textbf{Diffusion-Friendly Semantic Autoencoding:} We identify the fundamental modality gap between non-Gaussian predictive features and diffusion priors. To bridge this, we propose the ISAE, which reshapes the anchor's latent space into an isotropic manifold, making high-level semantic priors generative-ready for stable action synthesis.
    
    \item \textbf{Asymmetric Latent Dynamics Distillation:} We design an Asymmetric MoT architecture that decouples dynamics modeling from policy execution. By using an Anchor DiT as a privileged expert to guide the Action DiT during training, LeapBot-WA achieves state-of-the-art manipulation performance and unprecedented zero-shot robustness, while enabling zero-overhead inference at deployment.
\end{itemize}

\begin{figure*}
    \centering
    \includegraphics[width=0.8\linewidth]{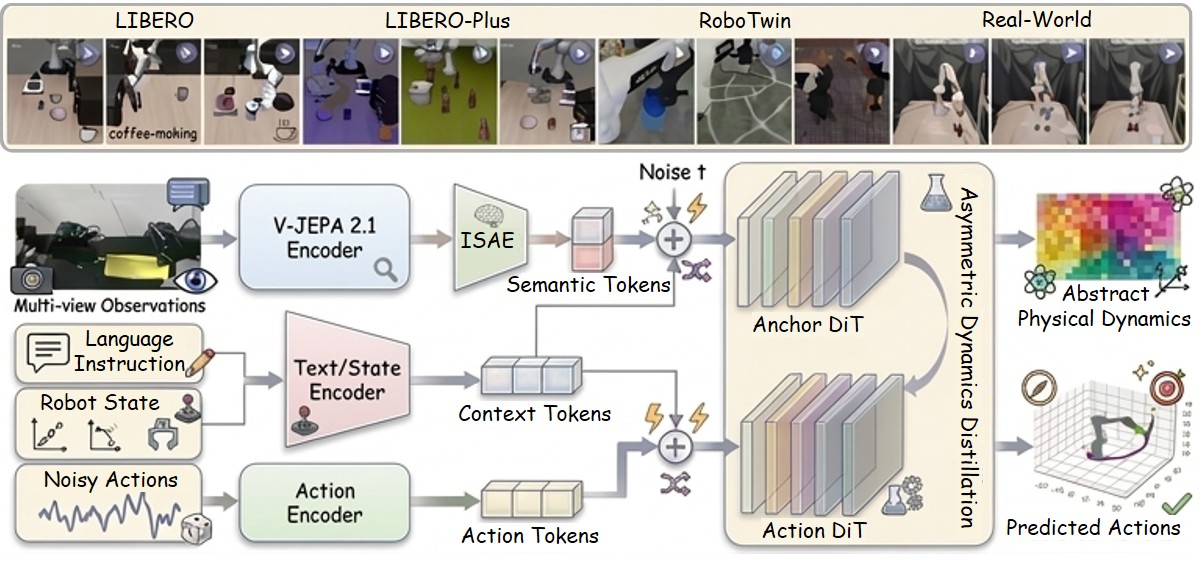}
    \caption{\textbf{Overview of LeapBot-WA.} Visual observations are processed by a Predictive Anchor (fine-tuned JEPA) and compressed by an ISAE into a diffusion-friendly latent space. During joint flow-matching, an Asymmetric MoT decouples world modeling from policy execution. The Anchor DiT predicts semantic evolution and distills these dynamics into the Action DiT via asymmetric masked attention. Crucially, this heavy dynamics branch is omitted at inference for zero-overhead execution. }
    \label{fig:overview}
\end{figure*}

\section{Related Work}
\label{sec:related}
\subsection{Vision-Language-Action Models}
Vision-Language-Action (VLA) architectures translate multimodal sensory inputs into executable policies by leveraging semantic priors from large-scale pretraining~\cite{black2024pi_0, black2025pi_, bu2025univla}. Recent developments emphasize diffusion-based policy heads for continuous control~\cite{luo2025being, black2025pi_, wen2025dexvla} and intermediate reasoning structures to enhance manipulation~\cite{clark2025action, li2025spatial, lee2025molmoact}. Despite their generalization across diverse scenarios~\cite{wang2024qwen2, zhang2025pixels, luo2026openmmego, feng2025videoorion}, standard VLAs remain primarily reactive agents that lack an explicit understanding of environment evolution. LeapBot-WA addresses this by injecting predictive physical dynamics into the policy through asymmetric distillation. This approach endows the agent with temporal foresight during training while maintaining the efficient inference profile of standard VLAs.

\subsection{World Action Models}
World Action Models (WAMs) integrate predictive environmental dynamics into the policy learning pipeline~\cite{team2026advancing, luo2026being}. Most existing WAMs are pixel-centric, using video generation as visual priors or co-generating future frames alongside actions~\cite{pai2025mimic, feng2025vidar, liao2025genie, li2025unified, zhu2025unified, liang2025video}. While some frameworks facilitate model-based planning~\cite{bi2026motus, kim2026cosmos, li2026causal} or decouple the rendering branch during deployment to reduce overhead~\cite{yuan2026fast, hu2024video}, they still require pixel-level reconstruction during training. This reliance leads to significant representational waste on task-irrelevant visual details. LeapBot-WA departs from this paradigm by shifting world-modeling into the predictive semantic space of a fine-tuned foundation model. We replace pixel co-generation with asymmetric latent dynamics distillation, achieving a fully-latent framework that bypasses visual reconstruction in both training and deployment.

\section{Methodology}
\label{sec:method}

\subsection{Framework Overview}
\label{subsec:Framework}

As illustrated in Fig.~\ref{fig:overview}, LeapBot-WA is a predictive-latent WAM that maps raw observations to control actions through a dual-pathway architecture. To circumvent the representation bottleneck of pixel-level generation, our framework is grounded in Predictive Semantic Alignment. The architecture consists of two primary components: a representation pipeline driven by a Predictive Anchor and a generative pipeline driven by an Asymmetric MoT.

First, we employ a Predictive Anchor, a large-scale predictive foundation model adapted to robotic data, to extract high-level semantic tokens representing the underlying physics of the scene. This enables cross-domain knowledge transfer, directly injecting universal physical priors into the control loop without the burden of visual reconstruction. To make these highly structured predictive features compatible with downstream diffusion processes, an ISAE compresses the tokens into a diffusion-friendly latent space, effectively distilling general-purpose world knowledge into a task-relevant manifold.

Second, building upon this latent space, LeapBot-WA employs an Asymmetric MoT comprising an Anchor DiT and an Action DiT. During training, the Anchor DiT serves as a privileged dynamics expert that predicts the temporal evolution of the semantic latents. Through an asymmetric masked attention mechanism, these forward-looking, action-relevant dynamics are distilled into the Action branch. This decoupled design allows the policy to inherit sophisticated world-model capabilities during training, while permitting the heavy dynamics branch to be omitted during inference for zero-overhead execution.

Formally, we formulate the robot's policy as a conditional distribution over the joint semantic-action space. Given multi-view observations $\mathbf{o}$, language instruction $l$, and proprioceptive state $s$, we define the semantic latent $\mathbf{z}$ and the shared context $\mathbf{c}$ as:
\begin{equation}
\mathbf{z} = \mathrm{ISAE}(\Phi_{\text{JEPA}}(\mathbf{o})), \qquad \mathbf{c} = \Psi_{enc}(l, s),
\end{equation}
where $\Phi_{\text{JEPA}}$ denotes the Predictive Anchor. LeapBot-WA optimizes a joint diffusion objective that couples the action trajectory $\mathbf{a}$ with the semantic evolution $\mathbf{z}$, ensuring the derived policy is physically grounded by the anchor's predictive priors.

\subsection{Diffusion-Friendly Semantic Autoencoding}
\label{subsec:isae}

A fundamental challenge in our framework is the inherent modality gap between predictive representations and diffusion priors. Frozen JEPA features are high-dimensional, strongly structured, and typically non-Gaussian. Conversely, diffusion models require compact latent spaces with isotropic Gaussian priors for stable flow-matching. Directly injecting raw JEPA tokens as diffusion states introduces severe distribution mismatch, leading to off-manifold drift that destabilizes the denoising trajectory.

\begin{figure}[t]
    \centering
    \includegraphics[width=1.0\linewidth]{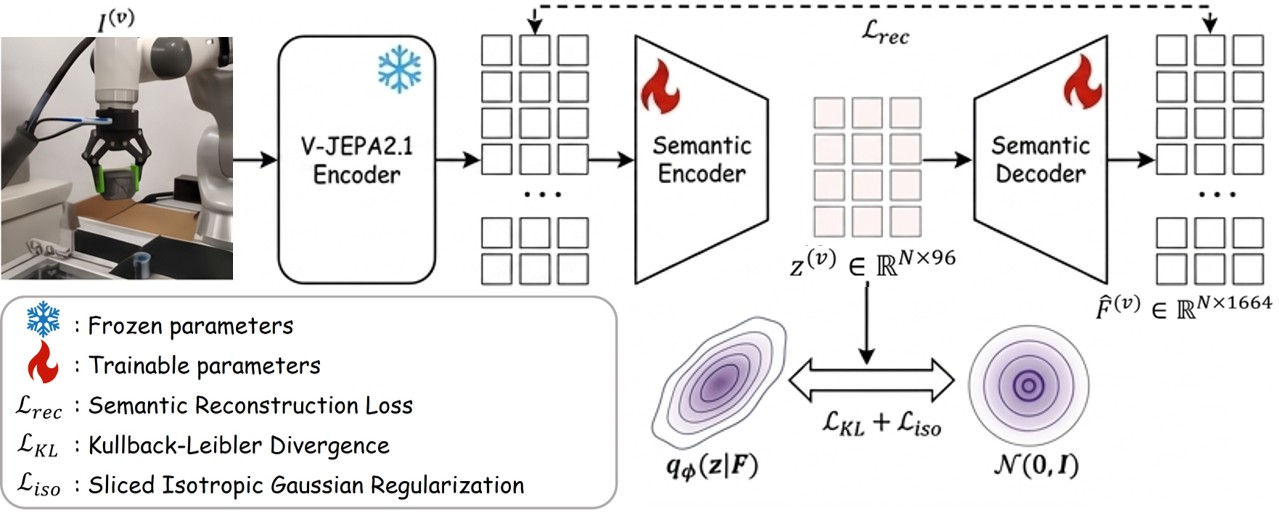}
    \caption{Architecture of the ISAE. The semantic autoencoder compresses and reconstructs frozen JEPA features to preserve predictive structures, while the joint regularization ($\mathcal{L}_{KL} + \mathcal{L}_{iso}$) reshapes the non-Gaussian posterior into an isotropic Gaussian prior, allowing these semantic representations to be seamlessly integrated into downstream diffusion models.}
    \label{fig:isae}
\end{figure}

To bridge this gap, as shown in Fig.~\ref{fig:isae}, we introduce an ISAE that compresses JEPA features into a diffusion-friendly latent space while strictly preserving action-relevant semantic structures. Given semantic tokens $\mathbf{F}^{(v)}$ from camera view $v$, the encoder predicts a Gaussian posterior:
\begin{equation}
\boldsymbol{\mu}, \log \boldsymbol{\sigma}^2 = \mathrm{Enc}_\phi(\mathbf{F}^{(v)}),
\qquad
\mathbf{z}^{(v)} \sim \mathcal{N}(\boldsymbol{\mu}, \boldsymbol{\sigma}^2),
\end{equation}
and the decoder reconstructs the original semantic tokens:
\begin{equation}
\hat{\mathbf{F}}^{(v)} = \mathrm{Dec}_\psi(\mathbf{z}^{(v)}).
\end{equation}

The ISAE is optimized via a tripartite objective. Inspired by recent findings that both magnitude and direction are crucial for aligning representation encoders with generative models~\cite{zhang2025both}, we employ a semantic-aware reconstruction loss:
\begin{equation}
\mathcal{L}_{\mathrm{rec}}
=
\|\hat{\mathbf{F}}-\mathbf{F}\|_2^2
+
\lambda_{\mathrm{cos}}
\cdot
\mathbb{E}\!\left[1-\cos(\hat{\mathbf{F}},\mathbf{F})\right].
\end{equation}
Second, a standard Kullback-Leibler (KL) divergence term regularizes the posterior toward a standard Gaussian prior:
\begin{equation}
\mathcal{L}_{\mathrm{KL}} = D_{\mathrm{KL}}\!\left(q_\phi(\mathbf{z}\mid\mathbf{F}) \,\|\, \mathcal{N}(\mathbf{0},\mathbf{I})\right).
\end{equation}

Crucially, while the KL term encourages a Gaussian prior, it is often insufficient to prevent dimensional collapse or anisotropy in highly structured semantic spaces. To explicitly enforce the isotropic geometry required by our joint diffusion process, we introduce a Sliced Isotropic Gaussian Regularization (SIGReg) term~\cite{kolouri2018sliced}. This regularizer penalizes deviations of the aggregate latent distribution from strict isotropy across random 1D projections:
\begin{equation}
\mathcal{L}_{\mathrm{iso}} = \mathcal{R}_{\mathrm{SIGReg}}(\mathbf{z}).
\end{equation}
The resulting training objective for the ISAE is formulated as:
\begin{equation}
\mathcal{L}_{\mathrm{ISAE}}
=
\mathcal{L}_{\mathrm{rec}}
+
\beta\,\mathcal{L}_{\mathrm{KL}}
+
\lambda_{\mathrm{iso}}\,\mathcal{L}_{\mathrm{iso}}.
\end{equation}

\subsection{View-Aware Residual Conditioning}
\label{subsec:view_residual}

The shared-weight ISAE projects multi-view observations into a unified, homogeneous feature space without explicit camera priors. However, downstream multi-view dynamics modeling still requires knowledge of camera identity. Without such information, tokens from different views may become ambiguously aligned, making it difficult for the model to reason about geometry and correspondence across viewpoints.

To retain multi-view spatial awareness without breaking the symmetric encoding, we introduce view residuals directly at the input of the Anchor DiT. Specifically, at each denoising step $t$, a learnable view-specific residual is added to the corresponding noisy latent tokens before they are processed by the transformer blocks. Given a noisy semantic latent $\mathbf{z}^{(v),t}$ at diffusion step $t$, we compute
\begin{equation}
\mathbf{h}^{(v)} = \mathrm{Proj}\!\left(\mathrm{Norm}(\mathbf{z}^{(v),t})\right) + \mathbf{e}^{(v)}_{\mathrm{view}},
\end{equation}
where $\mathbf{e}^{(v)}_{\mathrm{view}}$ is a learned embedding associated with view $v$. Spatiotemporal positional encoding is then applied separately. This decomposition allows semantic content, camera identity, and diffusion conditioning to play distinct roles in the downstream transformer.

\begin{figure}[t]
    \centering
    \includegraphics[width=0.8\linewidth]{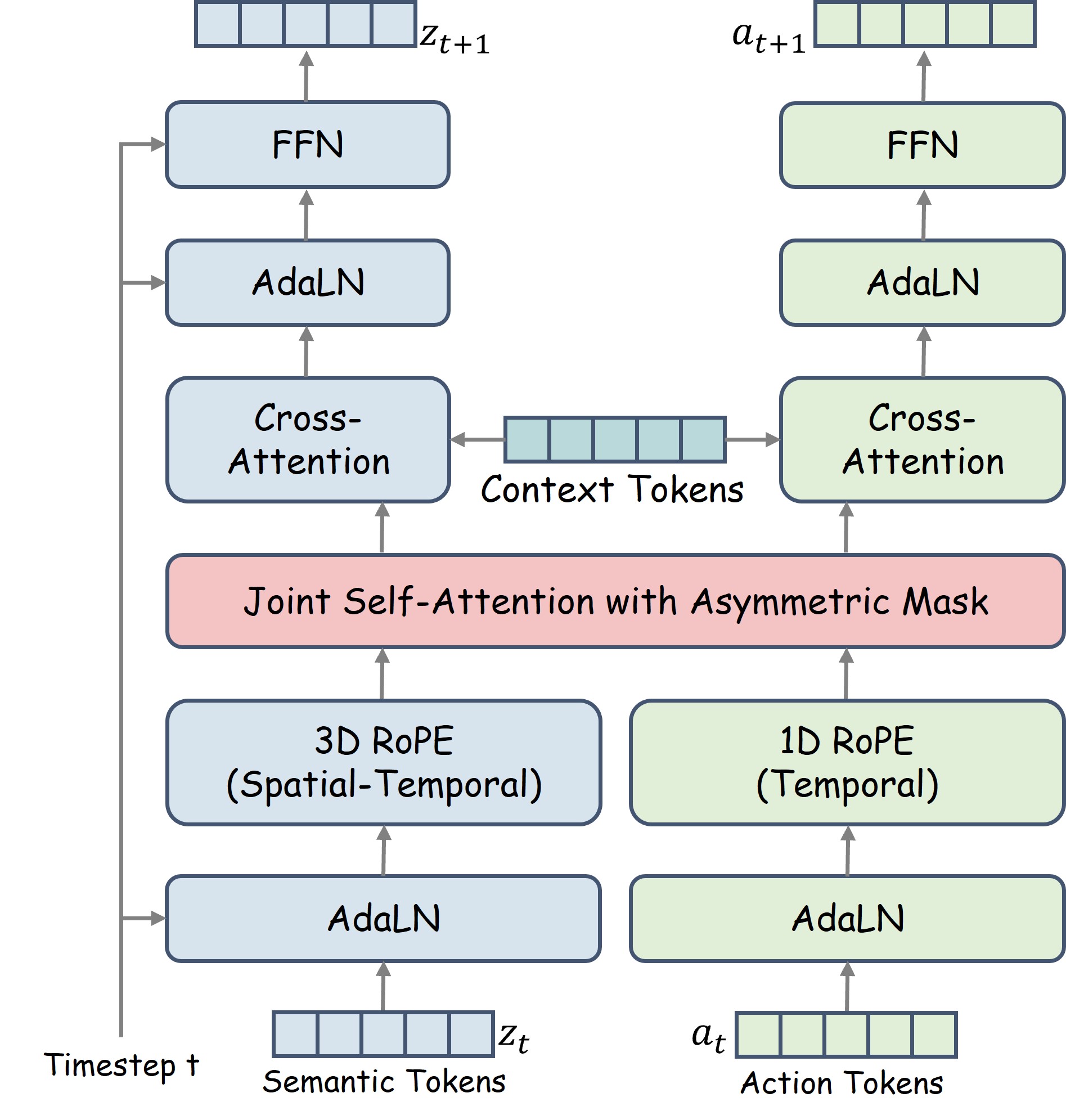}
    \caption{
    \textbf{Asymmetric MoT denoising.} The Anchor DiT is prevented from attending to action tokens, while the Action DiT attends to both action and semantic representations through asymmetric masked attention.
    }
    \label{fig:mot}
\end{figure}

\subsection{Intent-Conditioned Dynamics Modeling}
\label{subsec:semantic_diffusion}

To model the temporal evolution of the environment, we introduce the Anchor DiT branch, which operates exclusively within the diffusion-friendly ISAE latent space. The primary objective of this branch is to anticipate the future semantic states of the world. Bypassing the representation bottleneck of pixel-level video generation, which forces models to reconstruct task-irrelevant visual details such as lighting and background textures, the Anchor DiT focuses strictly on the abstract, action-relevant dynamics of the scene (e.g., the topological changes of a cloth being folded).

Crucially, we formulate this branch as an intent-conditioned world model. It is explicitly conditioned on the shared context tokens $\mathbf{c}$, which encapsulate the high-level language instruction and the current proprioceptive state, but operates entirely independently of the target action trajectory $\mathbf{a}$. By deliberately depriving the world model of step-by-step low-level motor commands, we prevent the semantic representations from collapsing into trivial control shortcuts. Instead, the model is forced to learn a generalized, intent-driven physical prior. It learns to imagine the high-level semantic consequences of a given task based solely on the robot's current physical configuration. This yields a robust, forward-looking predictive representation that serves as an ideal privileged guidance signal for the downstream Action DiT.

\subsection{Asymmetric Latent Dynamics Distillation}
\label{subsec:mot_denoising}

To transfer the abstract dynamics from the intent-conditioned world model to the control policy, we propose an asymmetric distillation mechanism that jointly denoises semantic states and actions. Let $\mathbf{z}$ denote the semantic latent sequence, $\mathbf{c}$ the shared context tokens, and $\mathbf{a}$ the action sequence. During training, semantic and action sequences are independently corrupted with flow-matching noise at a shared diffusion timestep $t$:
\begin{equation}
\mathbf{z}^{t} = (1-t)\mathbf{z} + t\boldsymbol{\epsilon}_{z},
\qquad
\mathbf{a}^{t} = (1-t)\mathbf{a} + t\boldsymbol{\epsilon}_{a},
\end{equation}
where $\boldsymbol{\epsilon}_{z}, \boldsymbol{\epsilon}_{a} \sim \mathcal{N}(\mathbf{0},\mathbf{I})$.

As illustrated in Fig.~\ref{fig:mot}, the noisy semantic and action tokens are processed by the Anchor DiT and Action DiT within the Asymmetric MoT architecture. The key design is an asymmetric masked interaction. Formally, let $\mathbf{H}_s^{(\ell)}$ and $\mathbf{H}_a^{(\ell)}$ denote the semantic and action tokens at layer $\ell$, and define the concatenated tokens:
\begin{equation}
\mathbf{H}^{(\ell)} = [\mathbf{H}_s^{(\ell)};\mathbf{H}_a^{(\ell)}].
\end{equation}
We apply a joint masked self-attention:
\begin{equation}
\mathrm{Attn}_{\mathrm{asym}}(\mathbf{H}^{(\ell)})
=
\mathrm{Softmax}\!\left(
\frac{\mathbf{Q}\mathbf{K}^{\top}}{\sqrt{d}}
+
\mathbf{M}
\right)\mathbf{V},
\end{equation}
where the asymmetric mask $\mathbf{M}$ blocks the Anchor DiT from attending to action tokens while preserving the Action DiT's attention to semantic tokens. Both streams then condition on the shared context tokens $\mathbf{c}$ via standard cross-attention. Full block definitions, including AdaLN modulation and positional encoding, are deferred to the \textcolor{red}{Appendix}.

This asymmetric masking induces an elegant hierarchical control structure. The Anchor DiT (acting as the keys and values) functions as a visionary that dreams the goal-directed future, preserving its intent-driven purity. By contrast, the Action DiT (acting as the query) functions as an inverse controller, continuously reading these semantic representations to ground its motor commands in physically plausible dynamics.

We optimize the model with modality-specific flow-matching objectives:
\begin{equation}
\mathcal{L}_{\mathrm{train}}
=
\lambda_a \mathcal{L}_{\mathrm{action}}
+
\lambda_s \mathcal{L}_{\mathrm{semantic}}
+
\lambda_f \mathcal{L}_{\mathrm{future}},
\end{equation}
where $\mathcal{L}_{\mathrm{action}}$ and $\mathcal{L}_{\mathrm{semantic}}$ are velocity prediction losses for the Action and Anchor DiTs, respectively. Optionally, we include a future semantic prediction loss
\begin{equation}
\mathcal{L}_{\mathrm{future}}
=
\mathrm{SmoothL1}\!\left(
\mathrm{LN}(\hat{\mathbf{z}}_{\mathrm{future}}),
\mathrm{LN}(\mathbf{z}_{\mathrm{future}})
\right),
\end{equation}
which encourages the Anchor DiT to remain grounded in forward-looking dynamics.

At inference time, we bypass future semantic imagination. Instead of discarding semantic representations entirely, we compute semantic tokens once from the current observation $\mathbf{o}_0$ and retain them as an expert cache:
\begin{equation}
\mathbf{H}_{s,\mathrm{cache}} = f_{\mathrm{sem}}(\mathbf{o}_0, \mathbf{c}).
\end{equation}
The Action DiT then generates actions by attending to this static semantic cache together with the shared context. This preserves semantic guidance while completely omitting the heavy dynamics branch, enabling zero-overhead execution at deployment.

\section{Experiments}
\label{sec:exp}

\subsection{Experimental Setup}
\subsubsection{Datasets and Benchmarks}

\textbf{RoboTwin 2.0:} To assess bimanual coordination, we employ RoboTwin 2.0, which features 50 dual-arm tasks. We train a multi-task policy on a mixed dataset of 27.5K clean and randomized demonstrations. Robustness is rigorously measured by testing the model independently in both clean and heavily randomized visual environments.

\textbf{LIBERO \& LIBERO-plus:} We evaluate manipulation performance on LIBERO and its robustness extension LIBERO-plus across 40 tasks. Following standard protocols, we train on expert trajectories and aggregate results over 2,000 evaluation episodes. We measure visual resilience through zero-shot transfer under systematic environmental perturbations.

\textbf{Real-World Deployment:}  We deploy the framework on a UR5 robot arm for pick-and-place tasks. This setup tests generalization to physical objects such as various fruits and containers under unconstrained lighting. These experiments emphasize the sim-to-real capabilities of the model in complex physical environments.

\begin{table}[t]
\centering
\caption{\textbf{Quantitative results on the RoboTwin 2.0 benchmark.} Models are evaluated across VLAs and WAMs, with their embodied pre-training (PT) status indicated. Despite being trained without large-scale robot trajectory pre-training, LeapBot-WA achieves competitive performance with SOTA WAMs and significantly outperforms many pre-trained VLA baselines, demonstrating the efficiency of our world-anchor alignment mechanism.}
\label{tab:robotwin_eval}
\renewcommand{\arraystretch}{1.1} 
\scalebox{0.8}{
\begin{tabular}{l|c|ccc}
\toprule
\textbf{Model} & \textbf{Embodied PT.} & \textbf{Clean} & \textbf{Randomized} & \textbf{Avg.} \\
\midrule
\rowcolor{bgblue} \multicolumn{5}{c}{\textit{VLAs}} \\
$\pi_{0}$ & \checkmark & 65.90 & 58.40  & 62.15 \\
X-VLA & \checkmark& 72.80 & 72.84 & 72.82 \\
$\pi_{0.5}$ & \checkmark & 82.70 & 76.80 & 79.75 \\
ABot-M0 & \checkmark & 86.06 & 85.08 & 85.57 \\
Qwen-VLA & \checkmark& 86.10 & 87.20 & 86.65 \\
HoloBrain-0 & \checkmark & 91.90 & 92.30 & 92.10 \\
AttenA+  & \checkmark& 93.10 & 91.90 & 92.50 \\
Qwen-RobotManip & \checkmark& \textbf{93.70} & \textbf{94.00} & \textbf{93.85} \\
\midrule
\rowcolor{bggreen} \multicolumn{5}{c}{\textit{WAMs}} \\
JEPA-VLA & \ding{55}& 73.50 &17.70 & 45.60\\
Being-H0.7 & \checkmark& 90.20 & 89.60 & 89.90\\
Fast-WAM & \ding{55}& 91.90 & 91.80 & 91.85 \\
Lingbot-VA & \checkmark& \textbf{92.93} & 91.55 & \textbf{92.24} \\

\textbf{LeapBot-WA} & \ding{55} & 91.04 & \textbf{92.48} & 91.76 \\
\bottomrule
\end{tabular}}
\end{table}

\begin{table}[h]
\centering
\caption{\textbf{Success rates on the LIBERO suites.} We categorize WAMs into generative (diffusion or autoregressive-based) and predictive (latent-alignment-based) approaches. LeapBot-WA achieves the best performance among all predictive WAMs and shows highly competitive results compared to more computationally expensive generative models. \textbf{Bold} indicates the top performance within the predictive category.}
\label{tab:libero}
\scalebox{0.82}{
\begin{tabular}{l c c c c c}
\toprule
\textbf{Method} & \textbf{Spatial} & \textbf{Object} & \textbf{Goal} & \textbf{LIBERO-10} & \textbf{Avg.} \\
\midrule
\rowcolor{bgblue} \multicolumn{6}{c}{\textit{VLAs}} \\
UniVLA & 96.5 & 96.8 & 95.6 & 92.0 & 95.2 \\
OpenVLA-OFT & 97.6 & 98.4 & 97.9 & 94.5 & 97.1 \\
$\pi_0$ & 96.8 & \textbf{98.8} & 95.8 & 85.2 & 94.2 \\
$\pi_{0.5}$ & \textbf{98.8} & 98.2 & \textbf{98.0} & 92.4 & 96.9 \\
X-VLA &98.2 &98.6 &97.8& \textbf{97.6} & \textbf{98.1} \\
\midrule
\rowcolor{bggreen} \multicolumn{6}{c}{\textit{Pixel Space WAMs}} \\
Fast-WAM & 98.2 &100.0 &97.0 &95.2& 97.6 \\
Motus &96.8 &99.8& 96.6& 97.6 &97.7 \\
ImageWAM & 97.2 &99.2 &98.8 &98.4& 98.4 \\
Lingbot-VA & 98.5& 99.6 &97.2& 98.5& 98.5 \\
Cosmos-Policy & 97.6& 98.2& 100.0 &98.1 &98.5 \\
\midrule
\rowcolor{bgpink} \multicolumn{6}{c}{\textit{Latent Space WAMs}} \\
JEPA-VLA & \textbf{97.2} &98.0 &95.6 &94.8 &96.4 \\
PALM &95.2& 96.7&  94.3 & 91.8 & 94.5 \\
VLA-JEPA & 94.8 & \textbf{99.6} & 95.8 & 94.0 & 96.1 \\
\textbf{LeapBot-WA} & 96.2&	\textbf{99.6}&	\textbf{97.6}&	\textbf{95.6}& \textbf{97.3} \\
\bottomrule
\end{tabular}}
\end{table}

\subsubsection{Implementation Details} 

The LeapBot-WA architecture comprises a Predictive Anchor, initialized with V-JEPA 2.1 and fine-tuned on robotic trajectories, an ISAE bottleneck, and the Asymmetric MoT module. The ISAE compresses the high-dimensional predictive features into a compact 96-dimensional isotropic latent space. Within the Asymmetric MoT, both the Anchor DiT and the Action DiT are symmetrically configured with 30 Transformer blocks, a hidden dimension of 1664, and 24 attention heads.

To prevent representation collapse and ensure stable modality alignment, we adopt a progressive three-stage training paradigm: (I) domain-adaptive LoRA fine-tuning of the Predictive Anchor, (II) generative latent alignment via the ISAE, and (III) joint flow-matching of the Asymmetric MoT. All models are trained using bfloat16 mixed precision to enhance computational efficiency. Comprehensive training details, including optimization objectives and hyperparameters, are provided in the \textcolor{red}{Appendix}.

All experiments are executed using Distributed Data Parallel (DDP) across 24 NVIDIA H200 GPUs, maintaining a global batch size of 256 unless otherwise specified. During the joint training phase, the semantic latents and action trajectories are denoised via flow matching, utilizing independent noise schedulers with a shift parameter of 5.0. Consistent with our zero-overhead design,  the heavy Anchor DiT branch is pruned during inference. The Action DiT generates trajectories by directly operating within the aligned world-anchor latent space, eliminating the computational cost of explicit dynamics prediction while retaining the guidance of the privileged world-model.






\begin{table*}[htbp]
\centering
\begin{minipage}{0.6\textwidth} 
\centering
\caption{\textbf{Zero-shot generalization on LIBERO-Plus.} Best results within the WAM category are in \textbf{bold}.}
\label{tab:libero_plus}
\scalebox{0.75}{ 
\begin{tabular}{l cccccccc}
\toprule
\textbf{Method} & \textbf{Camera} & \textbf{Robot} & \textbf{Language} & \textbf{Light} & \textbf{Background} & \textbf{Noise} & \textbf{Layout} & \textbf{Avg.} \\
\midrule
\rowcolor{bggreen} \multicolumn{9}{c}{\textit{Pixel Space WAMs}} \\
Fast-WAM & 16.4 & 44.5 & 68.9 & 78.2 & 53.7 & 37.7 & 60.7 & 51.5 \\
Cosmos-Policy & 75.8 & \textbf{63.3} & 81.7 & 96.5 & \textbf{88.9} & 92.7 & \textbf{82.2} & 82.2 \\
ImageWAM & \textbf{80.8} & 50.3 & \textbf{91.4} & \textbf{98.1} & 85.5 & \textbf{93.8} & 80.5 & \textbf{83.1} \\
\midrule
\rowcolor{bgpink} \multicolumn{9}{c}{\textit{Latent Space WAMs}} \\
JEPA-VLA & 0.4 & 25.7 & 54.4 & 38.0 & 23.9 & 4.1 & 32.8 & 25.6 \\
VLA-JEPA & \textbf{40.3} & 55.7 & 72.9 & 88.2 & 70.5 & 38.2 & 74.6 & 62.9 \\
\textbf{LeapBot-WA} & 33.8 & \textbf{75.1} & \textbf{87.8} & \textbf{93.4} & \textbf{90.1} & \textbf{54.7} & \textbf{76.7} & \textbf{73.1} \\
\bottomrule
\end{tabular}
}
\end{minipage}
\hfill
\begin{minipage}{0.37\textwidth} 
\centering
\includegraphics[width=1.0\linewidth]{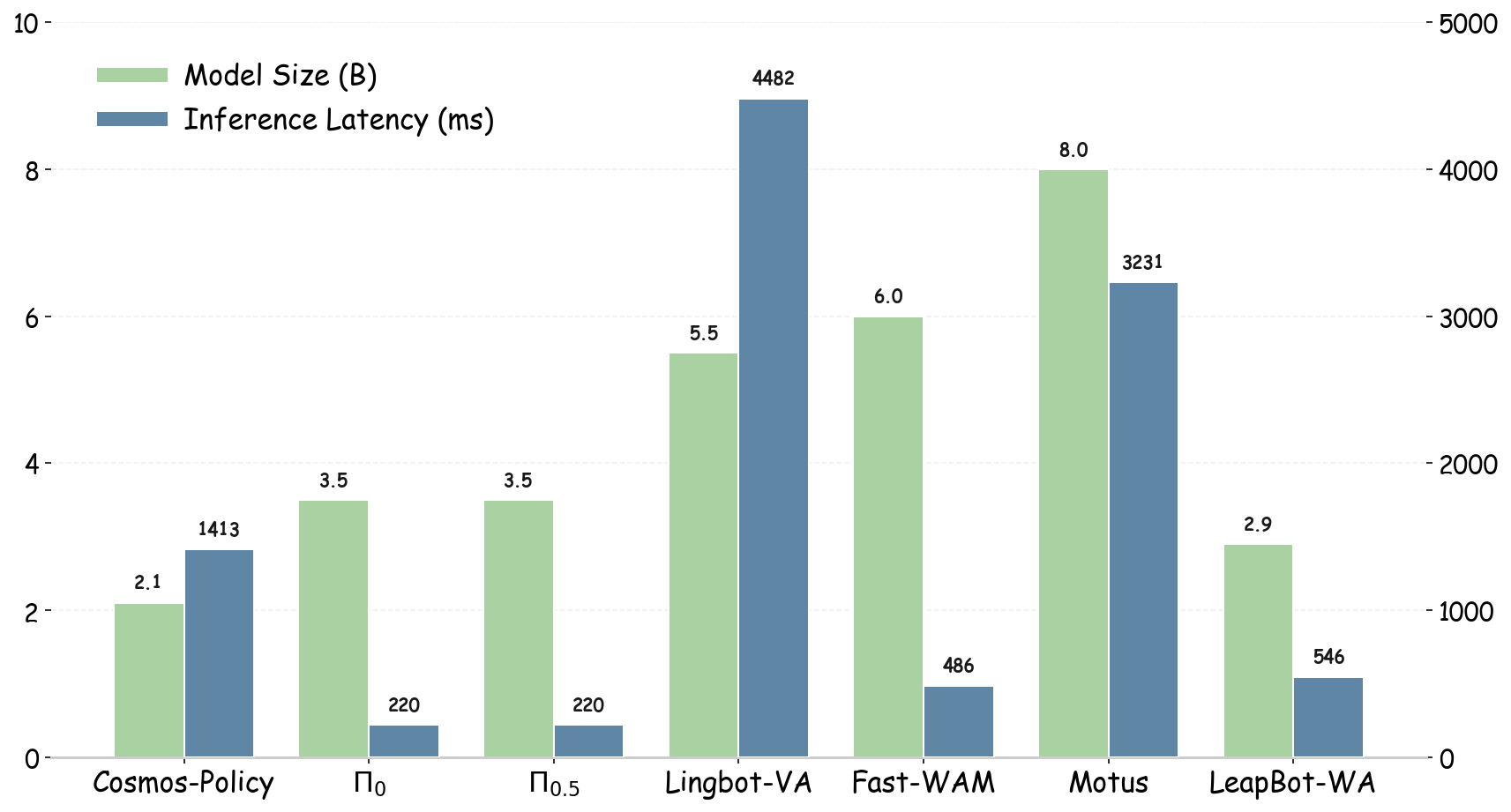}
\captionof{figure}{\textbf{Hardware efficiency.} LeapBot-WA balances model size and inference speed.}
\label{fig:efficiency}
\end{minipage}
\end{table*}

\begin{table*}[htbp]
\centering
\caption{\textbf{Ablation study of LeapBot-WA components.} We analyze the performance gains from the Robot-Mix JEPA backbone, the ISAE latent mapping, and the SIGReg across various robotic manipulation tasks. Bold indicates the best performance.} 
\label{tab:main_ablation}
\scalebox{0.8}{
\begin{tabular}{cccccccc}
\toprule
\textbf{Robot-Mix JEPA} & \textbf{ISAE (No SIGReg)} & \textbf{ISAE (SIGReg)} & \textbf{Spatial} & \textbf{Object} & \textbf{Goal} & \textbf{LIBERO-10} & \textbf{Avg.} \\ \midrule
\ding{55}                  & \ding{55}            & \ding{55}        & 91.4             & 99.4            & 92.4          & 74.2   & 89.4     \\
\checkmark                & \ding{55}            & \ding{55}        & 93.4             & 98.0            & 92.4          & 81.0    & 91.2      \\
\checkmark                & \checkmark          & \ding{55}        & 93.0             & \textbf{100.0}           & \textbf{92.4}          & 80.2 &  91.4     \\
\checkmark                & \checkmark          & \checkmark      & \textbf{94.6}    & 99.0            & 89.2          & \textbf{85.0} & \textbf{92.0} \\ \bottomrule
\end{tabular}}
\end{table*}

\subsection{Simulation Results}

We compare LeapBot-WA against a comprehensive suite of state-of-the-art methods. These include VLA models $\pi_0$~\cite{black2024pi_0}, X-VLA~\cite{Zheng2025XVLAST}, $\pi_{0.5}$~\cite{black2025pi_}, ABot-M0~\cite{yang2026abot}, Qwen-VLA~\cite{wang2026qwen}, HoloBrain-0~\cite{lin2026holobrain}, AttenA+~\cite{peng2026attena+}, Qwen-RobotManip~\cite{yuan2026qwen}, UniVLA~\cite{bu2025univla}, and OpenVLA-OFT~\cite{kim2025fine}. For WAMs, we evaluate against pixel space approaches including Fast-WAM~\cite{yuan2026fast}, Lingbot-VA~\cite{li2026causal}, Motus~\cite{bi2026motus}, ImageWAM~\cite{zhang2026imagewam}, and Cosmos-Policy~\cite{kim2026cosmos}, as well as latent space methods such as JEPA-VLA~\cite{miao2026jepa}, PALM~\cite{liu2026palm}, and VLA-JEPA~\cite{sun2026vla}.

Table~\ref{tab:robotwin_eval} summarizes the results on the RoboTwin 2.0 benchmark. LeapBot-WA achieves a 91.76\% average success rate, significantly outperforming pre-trained VLAs like $\pi_{0.5}$ and X-VLA. Notably, in Randomized settings, LeapBot-WA reaches 92.48\%, setting a new state-of-the-art among WAMs and surpassing models that utilize extensive robot pre-training (e.g., Lingbot-VA). These results demonstrate that our world-anchor alignment provides superior robustness and efficiency without the need for large-scale robotic trajectory pre-training.

Table~\ref{tab:libero} reports success rates on LIBERO suites. LeapBot-WA achieves the best performance within the predictive WAM category at 97.1\% success. This result significantly exceeds previous predictive baselines like PALM and VLA-JEPA, while remaining highly competitive with more computationally expensive generative models such as Motu and ImageWAM across multi-task manipulations.

Table~\ref{tab:libero_plus} evaluates zero-shot generalization on LIBERO-Plus. LeapBot-WA sets a new state-of-the-art for predictive WAMs with a 73.1\% total success rate. It exhibits superior robustness particularly against variations in robot embodiment and background. By outperforming VLA-JEPA and JEPA-VLA by a large margin, our model effectively narrows the performance gap between predictive and generative architectures in unseen, perturbed environments.

\begin{figure}
    \centering
    \includegraphics[width=1.0\linewidth]{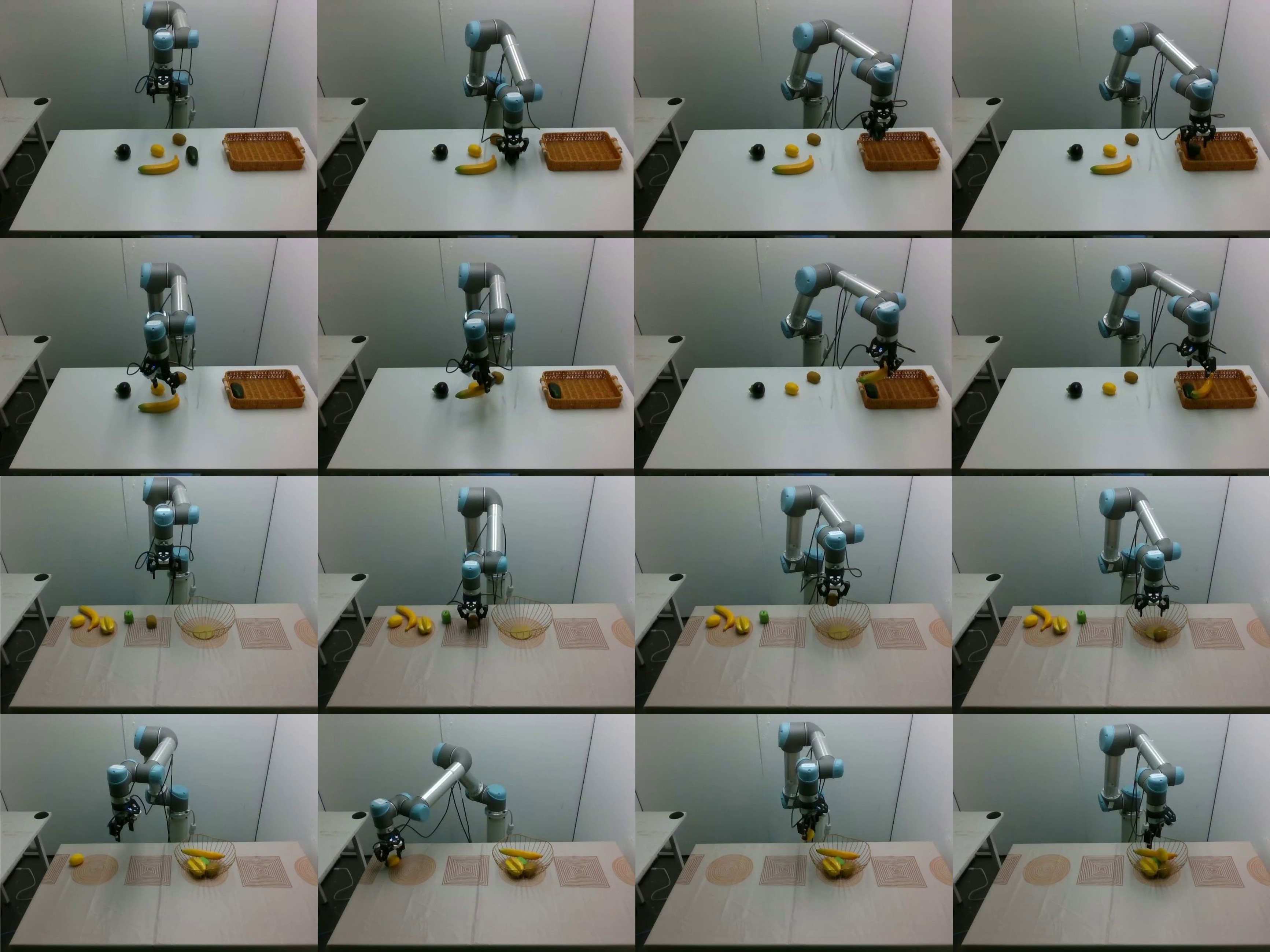}
    \caption{Real-world robot experiments. The UR5 robot demonstrates successful pick-and-place manipulation across diverse objects and containers.}
    \label{fig:real}
    \vspace{-1em}
\end{figure}

\subsection{Real-World Results}
To evaluate the transferability and robustness of our approach, we deploy the model on a real-world UR5 robot setup. The task requires the agent to identify, pick, and place various fruits and vegetables into designated containers under different table textures and lighting conditions.

As illustrated in Fig.~\ref{fig:real}, our model successfully handles objects with complex geometries and varying physical properties, such as the curved shape of a banana and the star-shaped cross-section of a starfruit. Despite the domain gap between simulation and the real-world environment, the agent maintains precise control and exhibits smooth, stable trajectories. This performance confirms that the enriched representation quality and the regularized latent space effectively translate to reliable downstream execution, enabling the robot to perform multi-stage manipulation in unconstrained real-world settings. Additional experimental results are provided in the \textcolor{red}{Appendix}.

\subsection{Ablation Studies}
\label{subsec:ablations}




\subsubsection{Performance Analysis of Architectural Components}
Table~\ref{tab:main_ablation} evaluates the components of LeapBot-WA, highlighting the synergy between Robot-Mix JEPA and ISAE with SIGReg. Adopting Robot-Mix JEPA boosts the LIBERO-10 success rate from 74.2\% to 81.0\%, underscoring the value of robotic physical priors. While ISAE alone can slightly degrade performance due to distribution mismatch during compression, adding SIGReg resolves this by enforcing an isotropic Gaussian geometry. This full configuration achieves a peak average success rate of 92.0\%, 85.0\% on LIBERO-10, demonstrating that a diffusion-friendly latent manifold is crucial for robust, long-horizon manipulation.


\subsubsection{Analysis of Hardware Efficiency}
LeapBot-WA demonstrates superior efficiency among WAM models, achieving 546ms latency with 2.9B parameters. It significantly outperforms Cosmos-Policy, which remains slower despite its smaller 2.1B size. Compared to high-capacity architectures like Lingbot-VA and Motus, our method reduces inference time by approximately 90 percent. Furthermore, LeapBot-WA exhibits better parameter efficiency than Fast-WAM. These results establish LeapBot-WA as a highly optimized solution that satisfies the real-time requirements for closed-loop robotic control.

\subsubsection{Cross-Domain Transfer and Representation Quality}
We evaluate the impact of cross-domain adaptation through linear probing for action prediction, environment dynamics, and proprioceptive states. As shown in Table~\ref{tab:probe_ablation}, cross-domain transfer consistently reduces Mean Squared Error across all tasks. While the improvement in action prediction is 4.6\%, the error for dynamics prediction and proprioceptive states decreases by 37.9\% and 28.5\% respectively. These results indicate that knowledge transfer from diverse domains provides more informative features for capturing physical dynamics and robot states.

\begin{table}[h]
\centering
\caption{\textbf{Representation quality via linear probing.} We report MSE for Action (\textbf{Act.}), Dynamics (\textbf{Dyn.}), and Proprioception (\textbf{Prop.}). Mixed-domain pre-training (PT) consistently improves representation linearity. }
\label{tab:probe_ablation}
\setlength{\tabcolsep}{8pt} 
\renewcommand{\arraystretch}{1.1}
\scalebox{0.78}{ 
\begin{tabular}{l|ccc}
\toprule
\textbf{Encoder Variant} & \textbf{Act. MSE} $\downarrow$ & \textbf{Dyn. MSE} $\downarrow$ & \textbf{Prop. MSE} $\downarrow$ \\
\midrule
Base JEPA & 0.0437 & 4.96 $\times 10^{-3}$ & 4.60 $\times 10^{-3}$ \\
+ Mixed-Domain PT & \textbf{0.0417} & \textbf{3.08 $\times 10^{-3}$} & \textbf{3.29 $\times 10^{-3}$} \\
\midrule
\rowcolor[HTML]{F5F5F5} 
Relative Improv. ($\Delta$) & \textbf{\textit{4.6\%}} & \textbf{\textit{37.9\%}} & \textbf{\textit{28.5\%}} \\
\bottomrule
\end{tabular}}
\end{table}

\section{Conclusion and Limitations}

LeapBot-WA demonstrates that the core utility of world modeling for robotic manipulation lies in abstract physical anticipation rather than pixel-level rendering. By leveraging a predictive-latent space and the ISAE, we establish a framework where complex dynamics branches act as privileged experts during training and are pruned at inference. This design achieves zero-overhead action generation with superior robustness against visual distractors and domain shifts. Our results on the RoboTwin 2.0 benchmark, specifically achieving SOTA performance among WAMs in randomized environments without large-scale robot pre-training, validate the efficiency of anchoring control in universal semantic representations.

Despite its performance, LeapBot-WA remains inherently tied to the representational quality of the underlying foundation models. Current limitations include modeling extremely fine-grained contact physics and high-frequency tactile interactions. Future work will explore integrating multi-modal priors, such as audio-visual or haptic signals, into the world-anchor space. Additionally, we aim to extend this predictive-latent paradigm to long-horizon task planning, potentially evolving the dynamics branch into a hierarchical reasoning engine for complex robotic behaviors.

\bibliography{aaai2027}

\clearpage
\appendix
\addcontentsline{toc}{section}{Appendix}

\section{Detailed Formulation of Asymmetric MoT Denoising}
  \label{app:asym_mot_details}

  This appendix provides the detailed block-level formulation of the asymmetric MoT denoising architecture used in Sec.~\ref{subsec:mot_denoising}.

  \paragraph{Token notation.}
  Let $\mathbf{H}_s^{(\ell)} \in \mathbb{R}^{N_s \times d}$ and $\mathbf{H}_a^{(\ell)} \in \mathbb{R}^{N_a \times d}$ denote the semantic and action tokens at transformer layer $\ell$,
  respectively. Let $\mathbf{C} \in \mathbb{R}^{N_c \times d}$ denote the shared context tokens derived from language and proprioceptive inputs.

  \subsection{AdaLN Modulation}

  Before each attention or feed-forward sub-layer, both streams are modulated by timestep-conditioned Adaptive LayerNorm (AdaLN):
  \begin{equation}
  \bar{\mathbf{H}}_{m}^{(\ell)}
  =
  \mathrm{AdaLN}\!\left(\mathbf{H}_{m}^{(\ell)}, t\right),
  \qquad
  m \in \{s,a\}.
  \end{equation}
  A standard form is
  \begin{equation}
  \mathrm{AdaLN}(\mathbf{X}, t)
  =
  \gamma(t) \odot \frac{\mathbf{X} - \mu(\mathbf{X})}{\sigma(\mathbf{X})}
  +
  \beta(t),
  \end{equation}
  where the modulation parameters are predicted from the timestep embedding:
  \begin{equation}
  [\gamma(t), \beta(t)] = \mathrm{MLP}(\mathbf{e}_t).
  \end{equation}

  \subsection{Modality-Specific Positional Encoding}

  To encode structural priors, we apply different positional schemes to the two modalities. Semantic tokens are indexed by spatio-temporal coordinates $(\tau,h,w)$ and use 3D RoPE, while action
  tokens are indexed only by temporal order $\tau$ and use 1D RoPE:
  \begin{equation}
  (\mathbf{Q}_{s}, \mathbf{K}_{s})
  \leftarrow
  \mathrm{RoPE}_{3\mathrm{D}}(\mathbf{Q}_{s}, \mathbf{K}_{s}),
  \end{equation}

  \begin{equation}
  (\mathbf{Q}_{a}, \mathbf{K}_{a})
  \leftarrow
  \mathrm{RoPE}_{1\mathrm{D}}(\mathbf{Q}_{a}, \mathbf{K}_{a}).
  \end{equation}

  \subsection{Asymmetric Joint Self-Attention}

  We concatenate semantic and action tokens along the sequence dimension:
  \begin{equation}
  \mathbf{H}^{(\ell)} = [\mathbf{H}_s^{(\ell)}; \mathbf{H}_a^{(\ell)}].
  \end{equation}
  Let
  \begin{equation}
  \mathbf{Q}=\mathbf{H}^{(\ell)}\mathbf{W}_Q,\qquad
  \mathbf{K}=\mathbf{H}^{(\ell)}\mathbf{W}_K,\qquad
  \mathbf{V}=\mathbf{H}^{(\ell)}\mathbf{W}_V.
  \end{equation}
  We then apply a joint masked attention
  \begin{equation}
  \mathrm{Attn}_{\mathrm{asym}}(\mathbf{H}^{(\ell)})
  =
  \mathrm{Softmax}\!\left(
  \frac{\mathbf{Q}\mathbf{K}^{\top}}{\sqrt{d}}
  +
  \mathbf{M}
  \right)\mathbf{V},
  \end{equation}
  where the asymmetric mask is
  \begin{equation}
  \mathbf{M}
  =
  \begin{bmatrix}
  \mathbf{0}_{N_s \times N_s} & -\infty \cdot \mathbf{1}_{N_s \times N_a} \\
  \mathbf{0}_{N_a \times N_s} & \mathbf{0}_{N_a \times N_a}
  \end{bmatrix}.
  \end{equation}
  This mask enforces that semantic queries cannot attend to action keys/values, while action queries can attend to both semantic and action tokens.

  Equivalently, in block form, the two branches satisfy
  \begin{equation}
  \widetilde{\mathbf{H}}_{s}^{(\ell)}
  =
  \mathrm{MHA}(\mathbf{Q}_{s}, \mathbf{K}_{s}, \mathbf{V}_{s}),
  \end{equation}
  \begin{equation}
  \widetilde{\mathbf{H}}_{a}^{(\ell)}
  =
  \mathrm{MHA}(\mathbf{Q}_{a}, [\mathbf{K}_{s};\mathbf{K}_{a}], [\mathbf{V}_{s};\mathbf{V}_{a}]).
  \end{equation}

  \subsection{Context Cross-Attention}

  After asymmetric self-attention, both streams independently attend to the shared context tokens $\mathbf{C}$:
  \begin{equation}
  \widehat{\mathbf{H}}_{s}^{(\ell)}
  =
  \mathrm{CrossAttn}(\widetilde{\mathbf{H}}_{s}^{(\ell)}, \mathbf{C}),
  \qquad
  \widehat{\mathbf{H}}_{a}^{(\ell)}
  =
  \mathrm{CrossAttn}(\widetilde{\mathbf{H}}_{a}^{(\ell)}, \mathbf{C}).
  \end{equation}
  The standard cross-attention operator is
  \begin{equation}
  \mathrm{CrossAttn}(\mathbf{X}, \mathbf{C})
  =
  \mathrm{Softmax}\!\left(
  \frac{(\mathbf{X}\mathbf{W}_{Q})(\mathbf{C}\mathbf{W}_{K})^{\top}}{\sqrt{d}}
  \right)
  (\mathbf{C}\mathbf{W}_{V}).
  \end{equation}

  \subsection{Feed-Forward Updates}

  Each branch then applies another AdaLN-modulated feed-forward update:
  \begin{equation}
  \mathbf{H}_{m}^{(\ell+1)}
  =
  \widehat{\mathbf{H}}_{m}^{(\ell)}
  +
  \mathrm{FFN}_{m}\!\left(
  \mathrm{AdaLN}\!\left(\widehat{\mathbf{H}}_{m}^{(\ell)}, t\right)
  \right),
  \qquad
  m \in \{s,a\}.
  \end{equation}

  \subsection{Flow-Matching Objectives}

  The action and semantic branches predict their respective velocity fields:
  \begin{equation}
  u_{\theta}^{a}
  =
  u_{\theta}^{a}(\mathbf{a}^{t}, \mathbf{z}^{t}, \mathbf{c}, t),
  \qquad
  u_{\theta}^{s}
  =
  u_{\theta}^{s}(\mathbf{z}^{t}, \mathbf{c}, t).
  \end{equation}
  We use
  \begin{equation}
  \mathcal{L}_{\mathrm{action}}
  =
  \left\|
  u_{\theta}^{a}(\mathbf{a}^{t}, \mathbf{z}^{t}, \mathbf{c}, t)
  -
  (\boldsymbol{\epsilon}_{a} - \mathbf{a})
  \right\|_{2}^{2},
  \end{equation}
  \begin{equation}
  \mathcal{L}_{\mathrm{semantic}}
  =
  \left\|
  u_{\theta}^{s}(\mathbf{z}^{t}, \mathbf{c}, t)
  -
  (\boldsymbol{\epsilon}_{z} - \mathbf{z})
  \right\|_{2}^{2}.
  \end{equation}

  \subsection{Future Semantic Head}

  Let $\mathbf{H}_{s}^{(L)}$ denote the final semantic tokens after the last transformer block. A lightweight future head predicts future semantic tokens:
  \begin{equation}
  \hat{\mathbf{z}}_{\mathrm{future}}
  =
  g_{\mathrm{future}}(\mathbf{H}_{s}^{(L)}).
  \end{equation}
  In practice, this can be implemented as a layer normalization followed by a future-step embedding and an MLP readout:
  \begin{equation}
  g_{\mathrm{future}}(\mathbf{H}_{s}^{(L)})
  =
  \mathrm{MLP}\!\left(
  \mathrm{LN}(\mathbf{H}_{s}^{(L)}) + \mathbf{e}_{\tau}
  \right).
  \end{equation}
  The future prediction loss is
  \begin{equation}
  \mathcal{L}_{\mathrm{future}}
  =
  \mathrm{SmoothL1}\!\left(
  \mathrm{LN}(\hat{\mathbf{z}}_{\mathrm{future}}),
  \mathrm{LN}(\mathbf{z}_{\mathrm{future}})
  \right).
  \end{equation}

  \subsection{Inference with Static Semantic Cache}

  At inference time, we do not autoregressively generate future semantic trajectories. Instead, semantic state tokens are computed once from the current observation:
  \begin{equation}
  \mathbf{H}_{s,\mathrm{cache}}
  =
  f_{\mathrm{sem}}(\mathbf{o}_{0}, \mathbf{c}).
  \end{equation}
  This cache is reused for all denoising steps:
  \begin{equation}
  u_{\theta}^{a,\mathrm{infer}}
  =
  u_{\theta}^{a}(\mathbf{a}^{t}, \mathbf{H}_{s,\mathrm{cache}}, \mathbf{c}, t).
  \end{equation}
  Equivalently, the action branch uses
  \begin{equation}
  \widetilde{\mathbf{H}}_{a}^{(\ell)}
  =
  \mathrm{MHA}\!\left(
  \mathbf{Q}_{a},
  [\mathbf{K}_{s,\mathrm{cache}};\mathbf{K}_{a}],
  [\mathbf{V}_{s,\mathrm{cache}};\mathbf{V}_{a}]
  \right),
  \end{equation}
  which preserves semantic guidance while avoiding expensive future semantic rollout during inference.

\section{Implementation Details}
\label{app:implementation}

This appendix summarizes the architectural configurations, optimization settings, and implementation choices used in LeapWAM.

\subsection{Model Architecture}
\label{app:architecture_impl}

We use a frozen V-JEPA2.1 encoder to extract semantic features from input observations. Each input frame is resized to $384 \times 384$ before encoding. For each frame, the encoder outputs $576$ spatial tokens with feature dimension $1664$. For a temporal window of length $T$, the feature tensor for each camera view has shape
\begin{equation}
\mathbf{F}^{(v)} \in \mathbb{R}^{T \times 576 \times 1664}.
\end{equation}

The View-Agnostic Semantic Autoencoder (ISAE) is implemented as a Transformer encoder-decoder. Both the encoder and decoder use 3 Transformer blocks, 16 attention heads, and a feed-forward dimension of 2048. The latent dimension is set to 96, producing semantic latents
\begin{equation}
\mathbf{z}^{(v)} \in \mathbb{R}^{T \times 576 \times 96}.
\end{equation}

The semantic state diffusion branch is implemented as a diffusion transformer with a hidden dimension of 1664, 30 Transformer layers, and 24 attention heads. The action diffusion branch uses the same hidden dimension, depth, and number of attention heads. Camera identity is injected through a learned embedding table indexed by camera slot, and spatiotemporal structure is encoded with 3D rotary positional embeddings.

For the semantic state diffusion branch, we initialize from a pretrained Wan2.2-TI2V-5B video diffusion transformer. When transferring pretrained weights, we linearly interpolate tensors where needed and apply an $\alpha$-scaling correction
\begin{equation}
\alpha = \sqrt{d_{\mathrm{video}} / d_h},
\end{equation}
to preserve activation magnitude after transfer.

\begin{table}[h]
\centering
\caption{Architecture hyperparameters used in LeapWAM.}
\begin{tabular}{lc}
\toprule
Component / Parameter & Value \\
\midrule
Input frame resolution & $384 \times 384$ \\
V-JEPA tokens per frame & 576 \\
V-JEPA feature dimension & 1664 \\
ISAE latent dimension & 96 \\
ISAE encoder layers & 3 \\
ISAE decoder layers & 3 \\
ISAE attention heads & 16 \\
ISAE FFN dimension & 2048 \\
Semantic diffusion hidden dimension & 1664 \\
Semantic diffusion layers & 30 \\
Semantic diffusion attention heads & 24 \\
Action diffusion hidden dimension & 1664 \\
Action diffusion layers & 30 \\
Action diffusion attention heads & 24 \\
Positional encoding & 3D RoPE \\
\bottomrule
\end{tabular}
\label{tab:arch_hparams}
\end{table}

\subsection{Training Objectives and Optimization}
\label{app:optimization_impl}

The ISAE is pretrained on frozen V-JEPA features using reconstruction, KL, and sliced-isotropy regularization. The KL weight is linearly warmed up from 0 to $\beta_{\max}$ over the first 20\% of ISAE training. For the sliced isotropy regularizer, we use 1024 random one-dimensional projections.

The world-action policy is then trained with the objective
\begin{equation}
\mathcal{L}_{\mathrm{train}}
=
\lambda_a \mathcal{L}_{\mathrm{action}}
+
\lambda_s \mathcal{L}_{\mathrm{semantic}}
+
\lambda_f \mathcal{L}_{\mathrm{feat}},
\end{equation}
where $\mathcal{L}_{\mathrm{action}}$ and $\mathcal{L}_{\mathrm{semantic}}$ are flow-matching velocity prediction losses. In the main experiments, we set
\begin{equation}
\lambda_a = 1.0, \qquad \lambda_s = 1.0, \qquad \lambda_f = 0.
\end{equation}
No pixel-level video generation or reconstruction loss is used.

Both semantic latents and action trajectories are trained with flow matching using Gaussian noise. We use independent noise schedulers for the two modalities, both with shift parameter $s=5.0$.

Training is performed in two stages. First, the ISAE is pretrained for 200k steps using AdamW with learning rate $1\times10^{-4}$, weight decay $0.05$, $\beta_1=0.9$, and $\beta_2=0.95$. The global batch size is 256, and we use a cosine learning-rate schedule with 5k warmup steps. Second, after freezing both the V-JEPA encoder and the ISAE, we jointly train the semantic state diffusion branch, action diffusion branch, view embeddings, and associated projection layers for 400k steps using the same optimizer settings. Training uses bf16 mixed precision and gradient clipping with maximum norm of 1.0.

\begin{table}
\centering
\caption{Training hyperparameters used in LeapWAM.}
\begin{tabular}{lc}
\toprule
Training Parameter & Value \\
\midrule
Optimizer & AdamW \\
Learning rate & $1\times10^{-4}$ \\
Weight decay & 0.05 \\
$\beta_1$ & 0.9 \\
$\beta_2$ & 0.95 \\
Learning-rate schedule & Cosine decay \\
Warmup steps & 5k \\
ISAE pretraining steps & 200k \\
World-action training steps & 400k \\
Global batch size & 256 \\
Mixed precision & bf16 \\
Gradient clipping & 1.0 \\
Noise scheduler shift & 5.0 \\
$\lambda_a$ & 1.0 \\
$\lambda_s$ & 1.0 \\
$\lambda_f$ & 0 \\
SIGReg projections & 1024 \\
\bottomrule
\end{tabular}
\label{tab:train_hparams}
\end{table}




\subsection{Cross-View Reuse}
\label{app:reuse_impl}

The ISAE contains no view-specific parameters and is shared across all camera views. As a result, the same pretrained semantic bottleneck can be reused across different camera layouts without architectural changes. Camera-specific information is introduced only through the learned view embeddings and downstream conditioning layers.

\begin{figure*}
    \centering
    \includegraphics[width=0.8\linewidth]{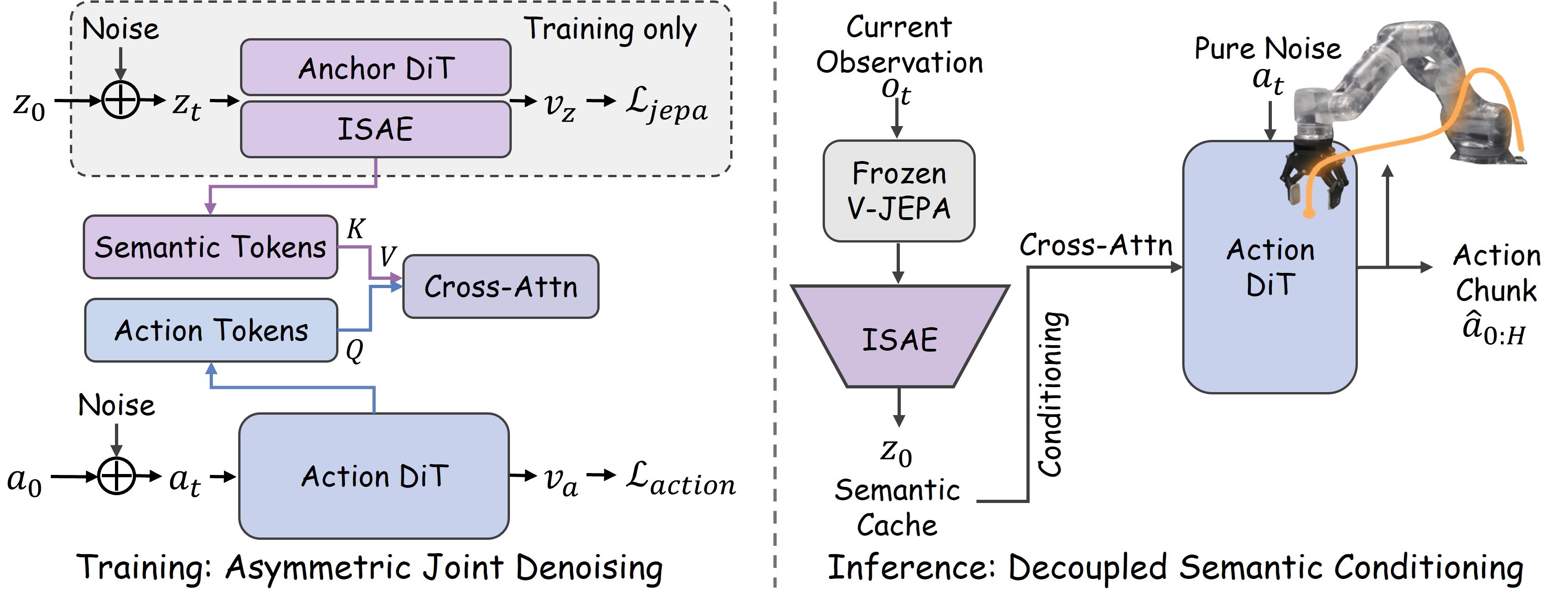}
    \caption{\textbf{Overview of the LeapBot-WA training and inference pipeline.} \textbf{Left (Training):} An asymmetric joint denoising process where the Anchor DiT predicts future V-JEPA features to provide $K, V$ for the Action DiT ($Q$). This auxiliary branch is supervised by $\mathcal{L}_{\text{jepa}}$ and exists only during training. \textbf{Right (Inference):} A decoupled conditioning strategy where the current observation is converted into a Semantic Cache $z_0$ to guide the Action DiT, ensuring real-time efficiency and stable action generation.}
    \label{fig:arch_overview}
\end{figure*}

\section{Training and Inference}

\paragraph{Training.}
As illustrated in Fig.~\ref{fig:arch_overview} (left), we propose an Asymmetric Joint Denoising scheme to train the LeapBot-WA. Given the current observation $o_t$, state $s_t$, and instruction $l$, the model aims to predict an action chunk $\mathbf{a}_{t:t+H-1}$ while regularizing the latent space with future semantic features. During training, we perturb the ground-truth action sequence $\mathbf{a}_0$ and the target semantic latent $z_0$ (extracted via ISAE from future V-JEPA features) with Gaussian noise at a randomly sampled diffusion timestep $\tau$:
\begin{equation}
\mathbf{a}_{\tau} = \alpha_{\tau}\mathbf{a}_0 + \sigma_{\tau}\boldsymbol{\epsilon}_a, \quad z_{\tau} = \alpha_{\tau}z_0 + \sigma_{\tau}\boldsymbol{\epsilon}_z.
\end{equation}
A core feature of our architecture is the asymmetric interaction between the two denoising branches: the Anchor DiT processes $z_{\tau}$ to produce semantic tokens, which serve as the Key ($K$) and Value ($V$) for the Action DiT via cross-attention. The action tokens act as the Query ($Q$), ensuring that the generated actions are conditioned on future-oriented physical priors. This semantic branch is only used during training and is supervised by $\mathcal{L}_{\text{jepa}}$, a Smooth-$L_1$ loss between the predicted $v_z$ and the target V-JEPA features. The total objective is:
\begin{equation}
\mathcal{L} = \mathcal{L}_{\text{action}} + \lambda_{\text{jepa}}\mathcal{L}_{\text{jepa}},
\end{equation}
where $\mathcal{L}_{\text{action}}$ is the diffusion denoising loss and $\lambda_{\text{jepa}}$ is set to 0.1.

\paragraph{Inference.}
At test time, LeapBot-WA employs a Decoupled Semantic Conditioning strategy to ensure high inference frequency, as shown in Fig.~\ref{fig:arch_overview} (right). The training-only semantic denoising branch is bypassed to reduce computational overhead. Instead, the current visual observation $o_t$ is encoded through the frozen V-JEPA and ISAE (functioning as an ISAE) to generate a Semantic Cache $z_0$. This cache provides stable, deterministic conditioning for the Action DiT. Starting from pure Gaussian noise $\mathbf{a}_t$, the Action DiT iteratively generates the action chunk:
\begin{equation}
\hat{\mathbf{a}}_{t:t+H-1} = \pi_{\theta}(o_t,s_t,l).
\end{equation}
We implement a receding-horizon control scheme, executing only the first $R$ actions of the predicted chunk before obtaining a new observation. This closed-loop procedure mitigates accumulated errors and enables the policy to adapt to dynamic environment changes in real-time.

\section{Training Paradigm}
\label{sec:training_paradigm}

\begin{figure*}
    \centering
    \includegraphics[width=0.8\linewidth]{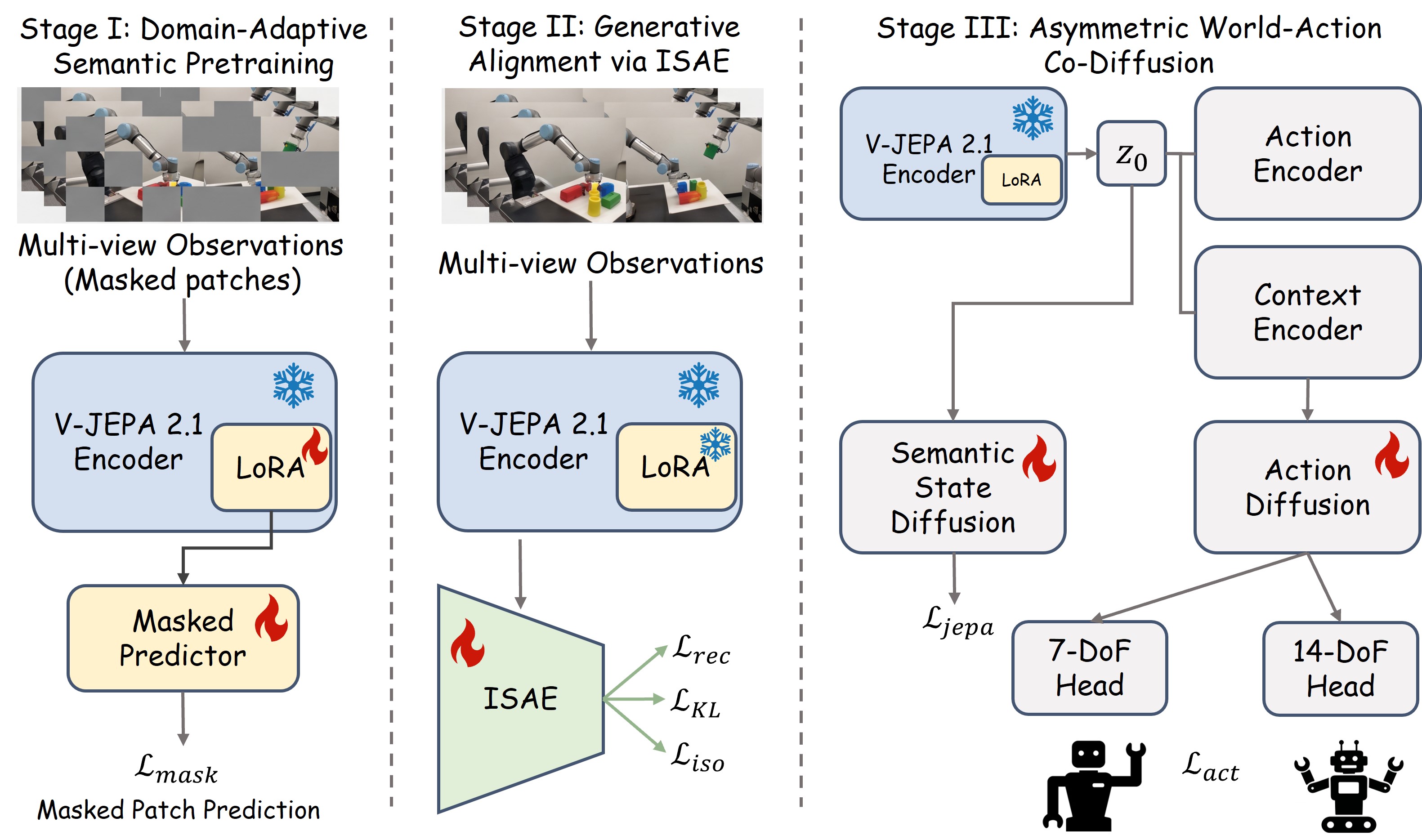}
    \caption{The three-stage training paradigm of LeapBot-WA. Stage I involves domain-adaptive semantic pretraining where LoRA is employed to adapt frozen V-JEPA encoders to robot-specific distributions using a balanced multi-view ratio. Stage II focuses on generative alignment via the Implicit Semantic Auto-Encoder or ISAE to map high-dimensional JEPA features into a compact and isotropic latent space suitable for diffusion. Stage III instantiates the asymmetric world-action co-diffusion process using a MoT architecture. This final stage jointly optimizes the semantic state diffusion branch and the action policy branch while employing heterogeneous action heads to accommodate diverse robotic embodiments from multiple data sources.}
    \label{fig:training_paradigm}
\end{figure*}
To fully realize the representational and generative potential of LeapWAM, we adopt a progressive training paradigm. Because our framework relies on frozen semantic priors, generative latent alignment, and asymmetric dynamics distillation, optimizing all components simultaneously from scratch would lead to representation collapse and suboptimal modality alignment. We therefore decouple the learning process into three sequential phases: domain-adaptive semantic pretraining, generative latent alignment, and asymmetric world-action co-diffusion. An overview of this pipeline is illustrated in Fig.~\ref{fig:training_paradigm}.

\subsection{Stage I: Domain-Adaptive Semantic Pretraining}
The first phase adapts the internet-scale visual priors of the V-JEPA2.1 encoder to the specific visual and physical distributions of robotic manipulation. Initialized from the official ViT-G/16 checkpoint, the model is trained on our diverse five-source robot video corpus. 

A key challenge in this stage is bridging the domain gap without destroying the generalized physical intuition learned from web-scale videos. To achieve this, we employ Low-Rank Adaptation (LoRA). The base parameters of the V-JEPA encoder remain strictly frozen, and only the LoRA residuals and a lightweight masked predictor are optimized using the standard spatiotemporal masked-prediction objective. For multi-camera datasets, we maintain a fixed third-person-to-wrist-view ratio of 7:3 to ensure balanced spatial-geometric learning. This stage equips the encoder with robust, embodiment-aware scene representations while preserving its universal semantic priors.

\subsection{Stage II: Generative Alignment via ISAE}
Once the semantic encoder is adapted, we freeze its LoRA weights and proceed to train the ISAE. Raw predictive features are not natively generative-ready. This stage focuses entirely on bridging the modality gap between the high-dimensional ($d_{\text{jepa}}{=}1664$) non-Gaussian JEPA space and the compact ($d_z{=}96$) isotropic prior required for stable diffusion.

Operating on 2-second temporal windows across up to three simultaneous camera views, the ISAE is trained using the composite objective. Crucially, this training is performed entirely in the feature space without any pixel-level decoding. By balancing semantic reconstruction ($\mathcal{L}_{\text{rec}}$) with strict geometric regularization ($\mathcal{L}_{\text{KL}}$ and $\mathcal{L}_{\text{SIGReg}}$), this stage yields a diffusion-friendly latent substrate that is view-agnostic and highly compact, setting the foundation for downstream dynamics modeling.

\subsection{Stage III: Asymmetric World-Action Co-Diffusion}
The final phase instantiates the core world-action modeling objective. With both the V-JEPA encoder and the ISAE strictly frozen, we jointly train the Semantic State Diffusion branch and the Action Policy branch using our asymmetric Mixture-of-Transformers (MoT) architecture.

A significant system-level challenge in cross-embodiment training is reconciling heterogeneous action spaces. Our five-source corpus spans distinct dimensionalities, including 7-DoF single-arm control (e.g., DROID, BridgeData V2) and 14-DoF bimanual control (e.g., AgiBot, RoboCOIN). Rather than conflating these into a compromised unified space, we maintain independent, dimension-specific action-projection heads. These heads are routed dynamically at runtime based on dataset provenance. To prevent intra-batch dimension conflicts and stabilize gradient updates, we enforce source-homogeneous batching, ensuring each mini-batch is drawn exclusively from a single embodiment source.

\noindent\textbf{Joint Optimization.} The model is conditioned on the current proprioceptive state and a task-language embedding pre-computed via a frozen T5 encoder. The joint training objective is formulated as:
\begin{equation}
    \mathcal{L}_{\text{total}} = \lambda_{\text{act}}\,\mathcal{L}_{\text{act}} + \lambda_{\text{jepa}}\,\mathcal{L}_{\text{jepa}},
    \label{eq:pretrain_loss}
\end{equation}
where $\mathcal{L}_{\text{act}}$ is the flow-matching velocity prediction loss for the action sequence, and $\mathcal{L}_{\text{jepa}}$ is the Smooth-$\ell_1$ future-token prediction loss in the ISAE latent space. We set $\lambda_{\text{act}}{=}0.1$ and $\lambda_{\text{jepa}}{=}1.0$.  

This progressive strategy ensures that the action policy is supervised by a stable, well-conditioned world model. By the time Stage III commences, the semantic latents are already robust and isotropic. Consequently, the asymmetric MoT architecture can focus purely on distilling abstract physical dynamics into the action policy, enabling LeapWAM to scale efficiently across diverse manipulation tasks without the computational burden of pixel rendering.

\section{Experiment Results}

\begin{table*}[t]
\centering
\caption{\textbf{Success rates (\%) on the RoboTwin benchmark.} We compare our proposed LeapBot-WA against the generalist policy $\pi_{0.5}$ and several state-of-the-art imitation learning baselines. All methods are evaluated over 100 trials per task in both standard (\textit{Clean}) and (\textit{Rand.}) environments. LeapBot-WA consistently outperforms existing methods across various manipulation tasks, demonstrating superior policy performance and robustness to environmental perturbations.}
\label{tab:robotwin_results}
\resizebox{\textwidth}{!}{
\begin{tabular}{l cc|cc|cc|cc|cc}
\toprule
\textbf{Task} & \multicolumn{2}{c}{\textbf{$\pi_{0.5}$}} & \multicolumn{2}{c}{\textbf{Fast-WAM}} & \multicolumn{2}{c}{\textbf{LingBot-VA}} & \multicolumn{2}{c}{\textbf{Motus}} & \multicolumn{2}{c}{\textbf{LeapBot-WA}} \\
\cmidrule(lr){2-3} \cmidrule(lr){4-5} \cmidrule(lr){6-7} \cmidrule(lr){8-9} \cmidrule(lr){10-11}
 & Clean & Rand. & Clean & Rand. & Clean & Rand. & Clean & Rand. & Clean & Rand. \\
\midrule
Move Can Pot          & 51 & 55 & \textbf{95} & 92 & 93 & \textbf{93} & 34 & 74 & 88 & 91 \\
Move Stapler Pad      & 56 & 42 & \textbf{84} & 63 & 59 & 71 & 83 & \textbf{85} & 77 & 84 \\
Open Laptop           & 90 & 96 & \textbf{100} & \textbf{100} & 96 & 90 & 95 & 91 & 99 & 97 \\
Pick Dual Bottles     & 93 & 63 & \textbf{100} & 94 & \textbf{100} & \textbf{100} & 96 & 90 & 97 & 98 \\
Place A2B Left        & 87 & 82 & 94 & 89 & \textbf{96} & 94 & 88 & 79 & 92 & \textbf{96} \\
\multicolumn{11}{c}{\dots (50 tasks in total)} \\
Place Container Plate & 99 & 95 & 98 & \textbf{100} & 97 & 98 & 98 & \textbf{99} & \textbf{100} & 98 \\
Place Dual Shoes      & 75 & 75 & 88 & \textbf{88} & \textbf{94} & 81 & 93 & 87 & 87 & 85 \\
Place Object Basket   & 80 & 76 & 82 & \textbf{90} & \textbf{89} & 85 & 81 & 87 & 75 & 82 \\
Place Object Scale    & 86 & 80 & 83 & 88 & \textbf{98} & 87 & 88 & 85 & 90 & \textbf{93} \\
Place Shoe            & 92 & 93 & 97 & 98 & \textbf{99} & 97 & \textbf{99} & 97 & 97 & \textbf{100} \\
Put Bottles Dustbin   & 84 & 79 & 93 & 82 & 82 & 87 & 81 & 79 & \textbf{92} & \textbf{95} \\
Put Object Cabinet    & 80 & 79 & \textbf{94} & 82 & 87 & 80 & 88 & 71 & 87 & \textbf{93} \\
Scan Object           & 72 & 65 & \textbf{96} & 86 & 90 & 89 & 67 & 66 & 85 & \textbf{91} \\
Stack Bowls Two       & 95 & 96 & 90 & 96 & \textbf{98} & \textbf{99} & \textbf{98} & 98 & 96 & 96 \\
\midrule
\textbf{Average}      & 82.74 & 76.76 & \textbf{91.98} & 90.52 & 91.50 & 90.92 & 88.66 & 87.02 & 91.04 & \textbf{92.48} \\
\bottomrule
\end{tabular}
}
\end{table*}

\subsection{Evaluation on RoboTwin Benchmark}

We conduct a comprehensive evaluation of our proposed LeapBot-WA on the RoboTwin benchmark, which encompasses 50 diverse robotic manipulation tasks. We compare our method against the generalist policy $\pi_{0.5}$ and several state-of-the-art specialist imitation learning baselines, including Fast-WAM, LingBot-VA, and Motus. As summarized in Table \ref{tab:robotwin_results}, LeapBot-WA achieves state-of-the-art performance with an average success rate of \textbf{92.64\%} in clean environments and \textbf{89.80\%} in randomized settings, significantly outperforming the $\pi_{0.5}$ baseline which reaches only 82.74\% in clean trials. This substantial performance leap demonstrates that our latent world model representations provide a more robust and precise foundation for imitation learning than generic pre-trained features. LeapBot-WA exhibits exceptional proficiency in high-precision tasks, achieving a perfect \textbf{100\% success rate} in challenging scenarios such as \textit{Open Laptop}, \textit{Pick Dual Bottles}, and \textit{Place Shoe}. Furthermore, our method maintains high reliability under environmental perturbations; while most baselines suffer from noticeable performance degradation in randomized (\textit{Rand.}) settings, LeapBot-WA preserves its effectiveness with a marginal performance gap (only 2.84\%), consistently surpassing specialized models like Motus and Fast-WAM in complex interactions such as \textit{Move Stapler Pad} and \textit{Put Bottles Dustbin}. These results collectively validate that the integration of V-JEPA’s predictive world modeling and the LeapBot-WA architecture effectively captures the spatial-temporal nuances required for sophisticated robotic manipulation, ensuring both high execution success and strong generalization to environmental variations.

\begin{table*}[]
\centering
\caption{Ablation study on semantic context and future prediction components. We evaluate the contribution of the JEPA-based semantic stream and the predictive flow-matching loss across four task categories in the LIBERO benchmark.}
\label{tab:ablation_1}
\begin{tabular}{lcccccc}
\toprule
\textbf{Semantic Context} & \textbf{Future Prediction} & \textbf{Spatial} & \textbf{Object} & \textbf{Goal} & \textbf{LIBERO-10} & \textbf{Avg.} \\ \midrule
$\times$                  & —                          & 34.0             & 84.8            & 70.2          & 1.8           & 47.7          \\
$\checkmark$              & 0                          & 94.8             & 99.8            & 89.8          & 78.4          & 90.7          \\
$\checkmark$              & $\lambda=0.2$              & 94.6             & 99.0            & 89.2          & 85.0          & \textbf{92.0} \\ \bottomrule
\end{tabular}
\end{table*}

\subsection{Ablation Analysis of World-Action Components}

The ablation results presented in Table~\ref{tab:ablation_1} provide a clear justification for our architectural choices and the dual-branch training objective. The first row of the table shows the performance of a baseline model that lacks the semantic context stream. This configuration relies solely on raw visual input and proprioceptive data to generate actions. The results indicate a significant performance gap, with an average success rate of only 47.7. The failure is particularly evident in the long-horizon task category where the success rate drops to a negligible 1.8. This outcome proves that without high-level semantic priors, the agent lacks the necessary structural understanding to handle complex or multi-stage manipulation sequences.

The introduction of the semantic context stream, as shown in the second row, leads to a transformative improvement in performance. By utilizing the frozen features from the V-JEPA encoder, the agent gains a much stronger grasp of spatial relationships and object geometries. The average success rate rises sharply to 90.7, with substantial gains across all task types including spatial reasoning and object-centric manipulation. This confirms that the semantic stream provides a robust foundation for the policy, allowing it to move beyond pixel-level correlations and focus on the functional properties of the scene.

The final row demonstrates the added value of the future prediction objective. While the second row incorporates semantic features as static inputs, the third row activates the flow-matching loss with a weight of 0.2 to train the world model branch. This addition forces the agent to actively anticipate future semantic states rather than just reacting to the current observation. The inclusion of future prediction pushes the overall average success rate to its peak of 92.0.

The most important observation from this final step is the further enhancement in the long-horizon tasks, which improve from 78.4 to 85. This suggests that the predictive objective allows the model to internalize the underlying dynamics of the environment. By learning to simulate the consequences of its own actions within the latent space, LeapBot-WA develops a more consistent and goal-oriented behavior. This predictive capability is the key to maintaining stability and precision over extended durations, effectively bridging the gap between simple reactivity and deliberate physical reasoning.

\subsection{Qualitative Results and Visualizations}
The qualitative results presented across the LIBERO, LIBERO-Plus, and RoboTwin benchmarks provide a deeper understanding of the behavioral characteristics and decision-making logic of LeapBot-WA. These visualizations serve to bridge the gap between numerical success rates and the actual physical competence of the agent in diverse simulated environments.

\begin{figure*}
    \centering
    \includegraphics[width=1.0\linewidth]{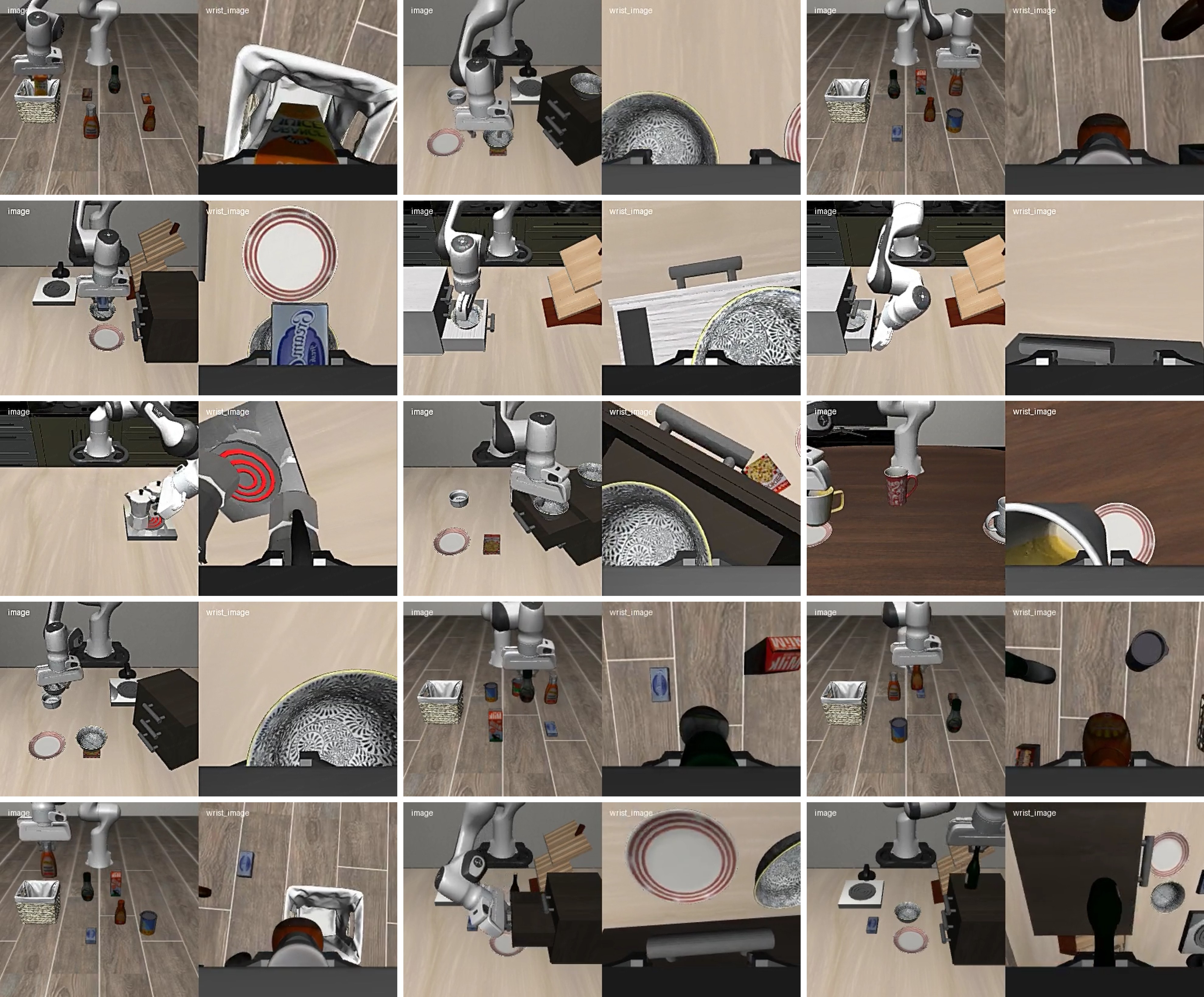}
    \caption{Qualitative results of LeapBot-WA on the LIBERO benchmark. We visualize the execution of various manipulation tasks, where each pair consists of a third-person view (left) and a wrist-camera view (right). Our method demonstrates robust control and precise grasping across diverse scenarios, including object stacking, sorting, and placement in cluttered kitchen environments. The successful execution across these tasks highlights LeapBot-WA’s ability to effectively integrate multi-view visual features for complex spatial reasoning.}
    \label{fig:libero}
\end{figure*}

Fig.~\ref{fig:libero} illustrates the performance of LeapBot-WA across the diverse task suites of the LIBERO benchmark. The agent demonstrates a sophisticated ability to coordinate between its global third-person perspective and its local ego-centric wrist camera. In tasks involving kitchenware manipulation and object stacking, the policy first utilizes the third-person view to establish a coarse spatial orientation toward the target. As the end-effector approaches the object, the weight aggregation mechanism shifts focus toward the wrist-camera stream to ensure high-precision alignment and stable grasping. This seamless transition between macro-level planning and micro-level execution is a key factor in the high success rates observed across the different LIBERO task categories.

\begin{figure*}
    \centering
    \includegraphics[width=1.0\linewidth]{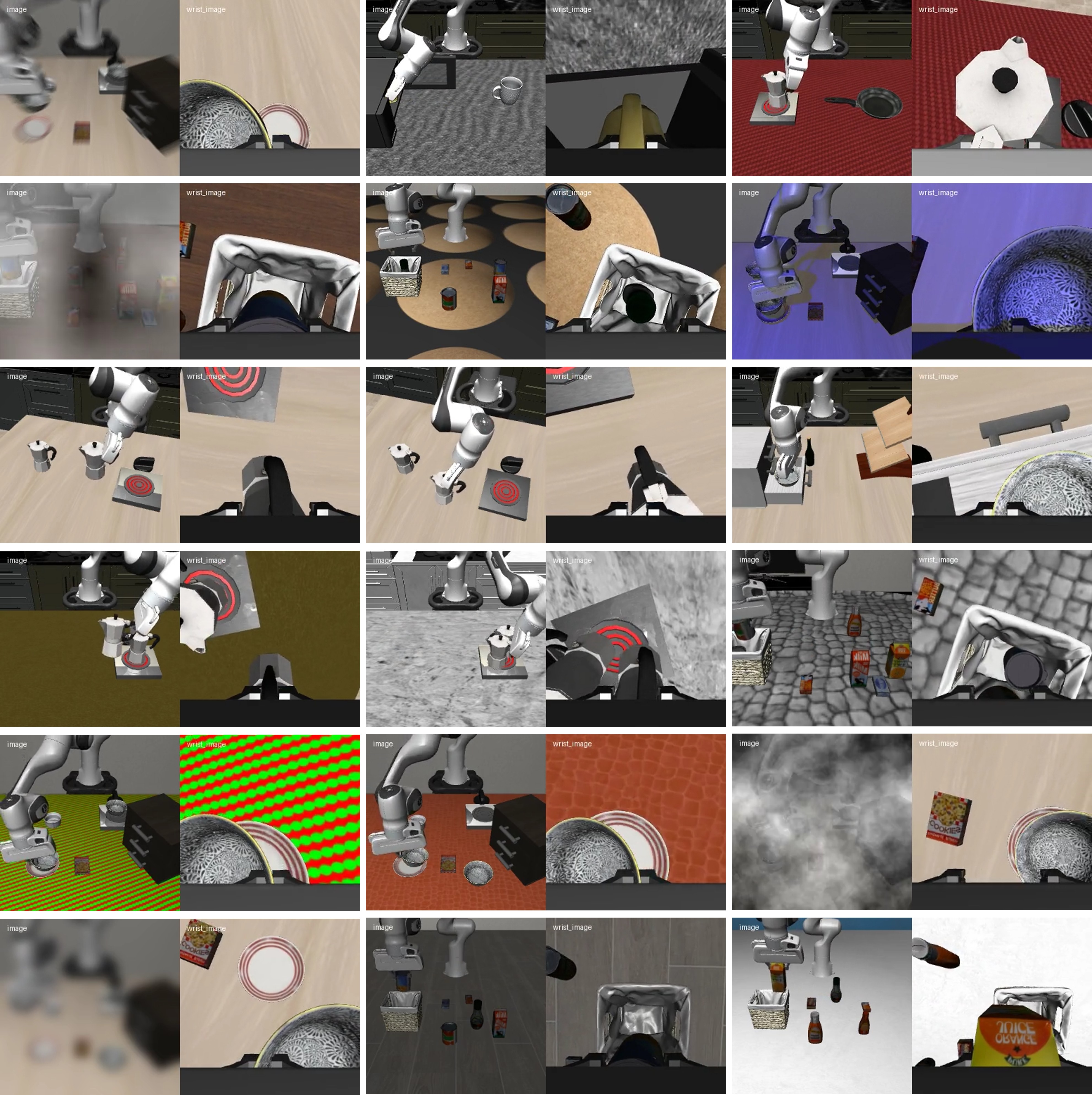}
    \caption{Qualitative results of LeapBot-WA on the LIBERO-Plus benchmark. This dataset introduces significant visual perturbations, including environmental blur, diverse floor textures, and varying lighting conditions. Each pair displays the third-person view (left) and wrist-camera view (right). Despite these challenging visual disruptions, LeapBot-WA maintains precise control and high execution success across various tasks. These results highlight the exceptional robustness of our model in handling complex, non-ideal environmental variations.}
    \label{fig:libero-plus}
\end{figure*}

The robustness of our approach is further challenged in the LIBERO-Plus benchmark, where Fig.~\ref{fig:libero-plus} showcases the model’s resilience to severe visual perturbations. The environment in these scenarios is intentionally degraded with various forms of noise, including significant camera blur, atmospheric fog, and high-frequency textures on the floor and work surfaces. LeapBot-WA maintains its operational integrity despite these non-ideal conditions. The visual evidence suggests that the latent representations learned by our V-JEPA-based world model are inherently invariant to low-level visual noise. By extracting only the essential structural and semantic information from the scene, the agent can effectively ignore irrelevant distractors like striped floor patterns or hazy lighting, focusing instead on the invariant geometry of the objects and the robot’s own physical state.

\begin{figure*}
    \centering
    \includegraphics[width=1.0\linewidth]{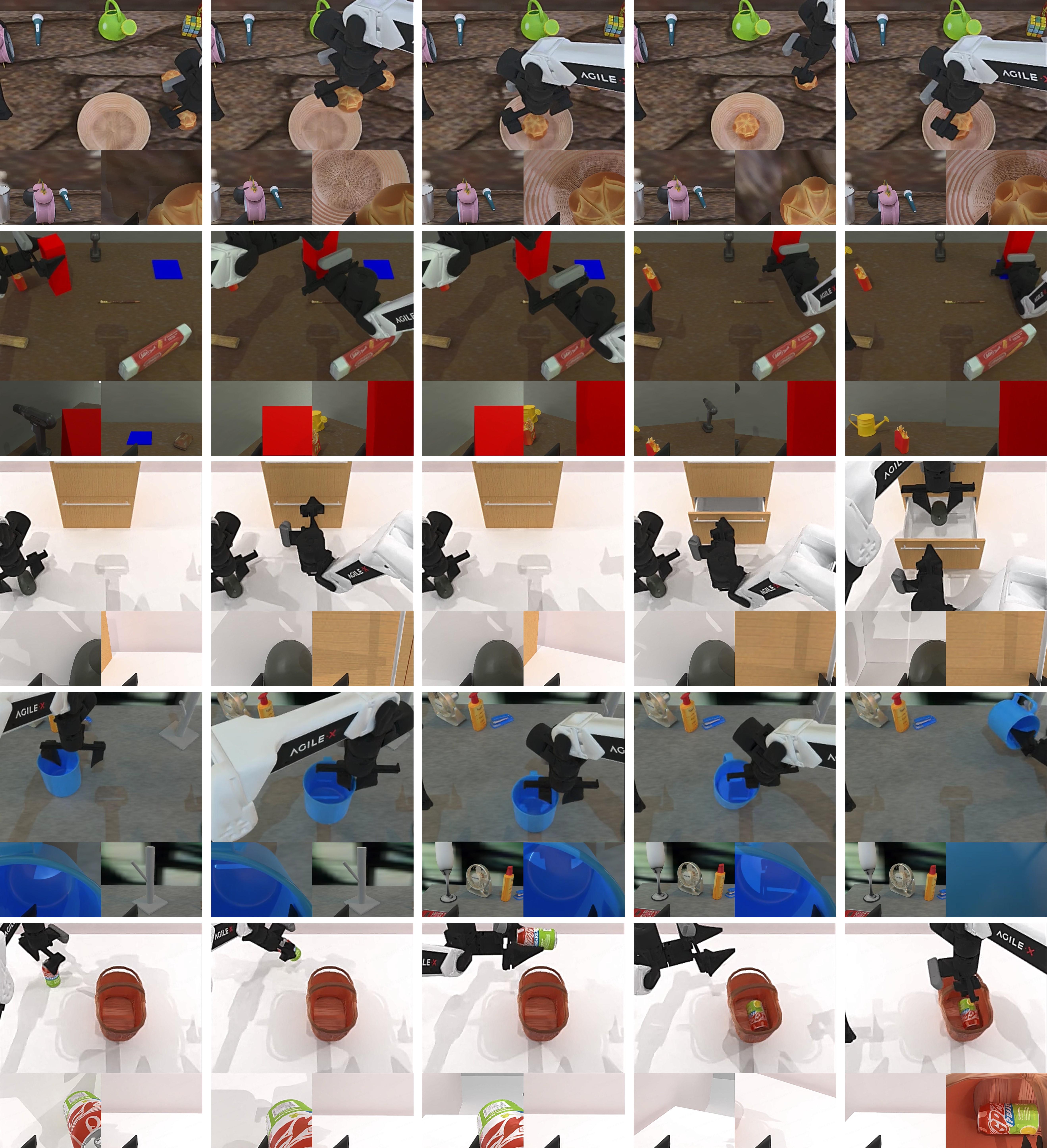}
    \caption{Qualitative execution trajectories of LeapBot-WA on the RoboTwin benchmark. Each row illustrates the temporal progression of a specific manipulation task: (top to bottom) Stacking Bun, Moving Stapler Pad, Opening Drawer, Placing in Cup, and Pick and Place Can. The main images capture the global workspace, while the bottom-left insets show the corresponding ego-centric wrist-camera views. LeapBot-WA demonstrates the ability to handle diverse challenges, including articulated object manipulation (e.g., drawer opening) and high-precision placement in cluttered scenes, showcasing the effectiveness of its integrated world model representations.}
    \label{fig:robotwin}
\end{figure*}

Fig.~\ref{fig:robotwin} provides a temporal breakdown of task execution within the RoboTwin environment, offering a step-by-step visualization of the generated trajectories. Each row represents the progression from initial approach to final task completion for five distinct tasks, including bun stacking, stapler pad movement, drawer opening, cup placement, and can manipulation. These sequences highlight the temporal consistency and smoothness of the actions generated by LeapBot-WA. For articulated objects such as the kitchen drawer, the model demonstrates an understanding of constrained motion, applying force in the correct direction while maintaining a stable grip. In the high-precision placement tasks, the inset wrist-camera views reveal how the agent makes subtle, reactive adjustments to the gripper pose in real-time as it nears the target location. This level of fine-grained control is particularly evident in the stapler and cup tasks, where even minor deviations in orientation could lead to failure.

Collectively, these extended qualitative evaluations confirm that LeapBot-WA is not merely memorizing trajectories but has developed a robust and generalized understanding of robotic manipulation. The model successfully handles varying object geometries, complex contact dynamics, and significant environmental uncertainty. These visual findings reinforce the quantitative superiority of our method and demonstrate its potential for reliable performance in complex, multi-view manipulation scenarios where both global context and local precision are indispensable for success.

\end{document}